\documentclass[11pt,a4paper]{article}
\usepackage{times,latexsym}
\usepackage{url}
\usepackage[T1]{fontenc}
\usepackage{tcolorbox}

\usepackage{enumitem}
\setlist{topsep=0pt, leftmargin=*}
\newlist{compactitem}{itemize}{3} % 3 is max-depth
\setlist[compactitem]{label=\textbullet, leftmargin=1em, labelindent=0.1em, itemsep=0em, parsep=0em}

\usepackage{amsmath}
\DeclareMathOperator*{\argmax}{\arg\!\max}

\usepackage[acceptedWithA]{tacl2021v1}
\usepackage{xspace,mfirstuc,tabulary}

\newif\iftaclinstructions
\taclinstructionsfalse % AUTHORS: do NOT set this to true
\iftaclinstructions
\renewcommand{\confidential}{}
\renewcommand{\anonsubtext}{(No author info supplied here, for consistency with
TACL-submission anonymization requirements)}
\newcommand{\instr}
\fi

\iftaclpubformat % this "if" is set by the choice of options

\else

\fi

\title{
\raisebox{1.4ex}{\includegraphics[width=0.8cm]{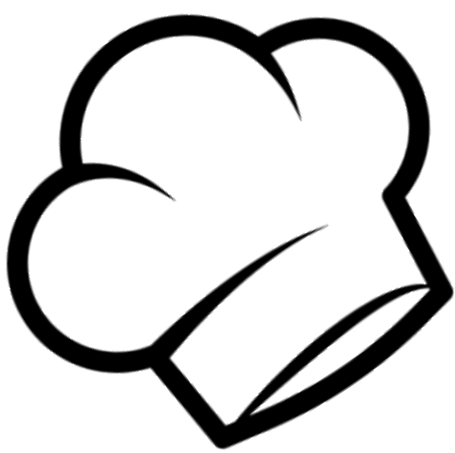}} 
\hspace{-0.35cm} 
$\boldsymbol{\mathcal{C}}$\textit{ooking}
\textit{Up}
$\boldsymbol{\mathcal{C}}$\textit{reativity}:
Enhancing LLM Creativity through \\Structured Recombination
}

\author{
    \textbf{Moran Mizrahi}$^\mathrm{1}$
    \;
    \textbf{Chen Shani}$^\mathrm{2}$
    \;
    \textbf{Gabriel Stanovsky}$^\mathrm{1}$
    \\
    \textbf{Dan Jurafsky}$^\mathrm{2}$
    \;
    \textbf{Dafna Shahaf}$^\mathrm{1}$
    \\%[0.5em]
    $^\mathrm{1}$The Hebrew University of Jerusalem\quad $^\mathrm{2}$Stanford University
    \\%[0.5em]
    \small{\texttt{\{moranmiz, gabis, dshahaf\}@cs.huji.ac.il\quad
    \{cshani, jurafsky\}@stanford.edu}}
    }
\date{}

\usepackage{graphicx}
\usepackage{booktabs}
\usepackage{multirow}
\usepackage{amsmath}
\usepackage[normalem]{ulem}
\usepackage{caption}
\usepackage{subcaption}
\usepackage{mathtools}
\usepackage{amsfonts}
\usepackage{longtable}
\usepackage{supertabular}
\usepackage{hyperref}
\usepackage[normalem]{ulem}
\usepackage{booktabs}
\usepackage{multirow}
\usepackage{cleveref}
\usepackage{changepage}
\usepackage{amsthm} 
\usepackage[mathscr]{euscript}
\usepackage{xcolor}

\newtheorem{definition}{Definition} 

\definecolor{blue-violet}{rgb}{0.54, 0.17, 0.89}
\definecolor{blue-violet2}{rgb}{0.9, 0.35, 0.89}
\definecolor{shamrockgreen}{rgb}{0.0, 0.62, 0.38}
\definecolor{psychedelicpurple}{rgb}{0.7, 0.0, 1.0}
\definecolor{azure}{rgb}{0.0, 0.5, 1.0}

\usepackage{tikz}
\usepackage{calc}

\usepackage{longtable}
\usepackage{booktabs}

\newcommand{\remove}[1]{} 

\newcommand{\xhdr}[1]{\vspace{1mm}\noindent{{\bf #1.}}} 
\newcommand{\xhdrnodot}[1]

\newcommand{\draftcomment}[3]{{\textcolor{#3}{[#1 -- #2]}}}

\newcommand{\mnote}[1]{\draftcomment{#1}{\textsc{moran}}{blue-violet2}} % tTODO: comment out
\newcommand{\cnote}[1]{\draftcomment{#1}{\textsc{Chen}}{red}}

\newcommand{\gabi}[1]{\draftcomment{#1}{\textsc{gabi}}{orange}} % TODO: comment out

\newcommand{\gabis}[1]{\gabi{#1}}

\newcommand{\alg}[0]{$\mathcal{D}$\textsl{ish}\textsc{Cover}}

\newcommand{\algbold}[0]{$\boldsymbol{\mathcal{D}}$\textbf{\textsl{ish}\textsc{Cover}}}

\newcommand{\gpt}[0]{GPT-4o}

\newcommand{\cgpt}[0]{GPT-4o}

\makeatletter
\newcommand\footnoteref[1]{\protected@xdef\@thefnmark{\ref{#1}}\@footnotemark}
\makeatother

\begin{document}
\maketitle

\begin{abstract}
Large Language Models (LLMs) excel at many tasks, yet they struggle to produce truly creative, diverse ideas.
In this paper, we introduce a novel approach that enhances LLM creativity. We 
apply LLMs for translating between natural language and structured representations, and perform the core creative leap via cognitively inspired manipulations on these representations.
Our notion of creativity goes beyond superficial token-level variations; rather, we recombine structured representations of existing ideas, enabling our system to effectively explore a more abstract landscape of ideas.

We demonstrate our approach in the culinary domain with \alg{}, a model that {generates creative recipes}.
Experiments and domain-expert evaluations reveal that our outputs, which are mostly coherent and feasible, significantly surpass \cgpt{} in terms of novelty and diversity, thus outperforming it in creative generation. 
We hope our work inspires further research into structured creativity in AI.

\end{abstract}

\section{Introduction}
\label{sec:intro}
\begin{figure*}[h]
    \centering
    \includegraphics[width=\textwidth]{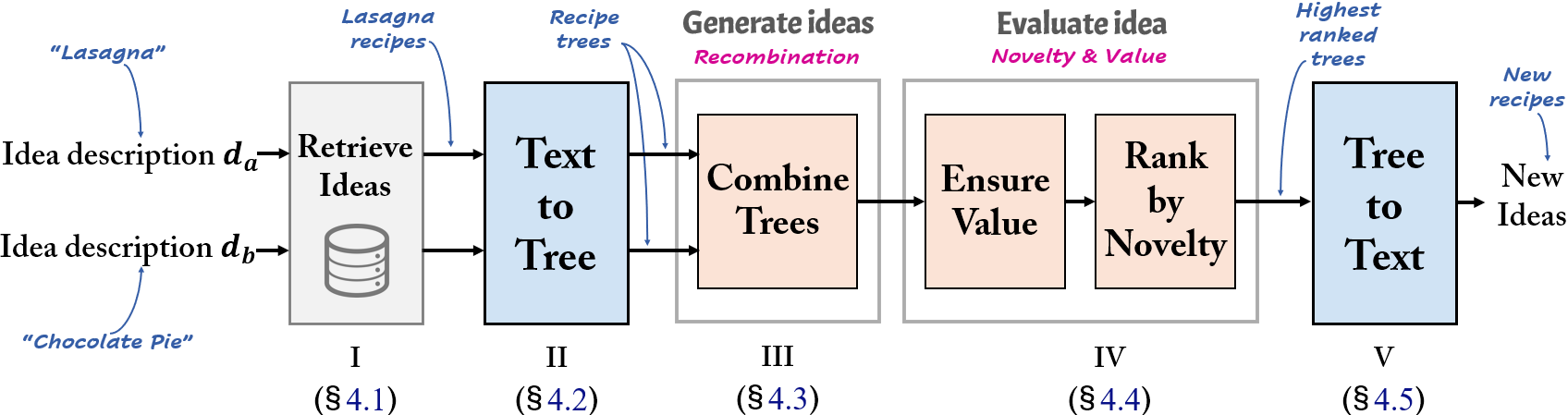}
    \caption{The \alg{} pipeline for creative recipe generation (with LLM-based components shaded in blue) takes as input two idea descriptions. Each 
    description 
    is mapped to a set of specific recipes (\S~\ref{algo:sampling}), which are parsed into tree representations (\S~\ref{algo:text_to_tree}). These trees are subsequently combined using a minimal edit distance algorithm (\S~\ref{algo:edit_distance}), 
    assessed for value, and ranked based on novelty scores 
    % their value is assessed, and they are ranked based on their novelty scores 
    (\S~\ref{algo:eval}). Finally, the highest-ranked trees are translated back into natural language recipes (\S~\ref{algo:tree_to_text}).
    % \gabis{Point to the different subsections from the Figure, e.g., under ``Combine Trees'' write ``\S4.3'' ( I think), and use the same titles for the subsection as the names that appear here, would make reading much easier.}
    }
    \label{fig:pipeline}
\end{figure*}

 %I think your idea is closely related to lots of recent literature on making LLMs better by combining  structured knowledge with generative LLM output (i'm thinking of things like https://arxiv.org/pdf/2407.06564 or maybe in combining parse information with LLMs: https://aclanthology.org/2024.acl-long.384/ but you can probably find better stuff. Your novelty is maybe showing that combining structured knowledge with LLMs surprisingly works also for creativity, something I wouldn't have guessed. Like the parse literature,  you generate the structured knowledge from the text itself rather than an outside knowledge source, but like the knowledge graph literature you dont' just parse but add something to the structure  (in your case recombination). I guess my advice is to go re-read that literature and figure out if there is some angle in applying this way of thinking to creativity specifically that has implications {\bf both} for the rest of LLM "combine LLM and structure" research and the "make LLMs more creative" research. And that might help you decide which parts of your method need to stay in the paper, and which, as Gabi and Chen suggest, could just be moved to the appendix as low-level implementaiton details

\remove{
 Our goal in this paper is to make LLMs more creative

Lots of work focuses on temperature, but that doesn't really work

 There is a lot of recent work showing that LLMs can get better with structured knowledge

 We manipulate the structure (recombination) and sample on this space

 ------------------
}

Large Language Models (LLMs) excel at generating fluent{, coherent} text and {performing} tasks that draw on extensive world knowledge. However, they often struggle to generate truly creative ideas
% , arguably due to their reliance on pattern learning
~\cite{franceschelli2024creativity, chakrabarty2024art, tian2024large, zhao2024assessing}. 

In creativity research, creative output{s} {are} typically defined as {those that are} both \textit{\textbf{novel}} (unexpected {and original}) and \textit{\textbf{valuable}} (useful{, relevant, or effective}) \cite{mumford2003have, boden2004creative}. 
However, due to LLM relying on vast repositories of existing data, they inherently follow learned patterns, making them prone to {producing} \textbf{predictable, repetitive} outputs that lack {genuine} novelty.
Ironically, attempts to {explicitly} instruct LLMs to ``think more creatively” {often} lead them to generate \textbf{invalid or hallucinated} (i.e., invaluable) solutions that could mislead uninformed users~\cite{wang2024preliminary, jiang2024survey}. 
Together, these limitations make creative generation a persistent challenge for LLMs.

The \emph{temperature} parameter of LLMs controls the amount of randomness, and is often claimed to be the creativity parameter, i.e., the implicit way to enhance creativity in LLM.
However, creativity {encompasses} much more than {mere} randomness; a recent study \cite{peeperkorn2024temperature} %evaluated this claim and
found that while {higher} temperature{s} weakly correlate with {increased} novelty, {their actual} influence on {overall} creativity {remains} {subtle and limited}. 

Much recent work has shown that combining LLMs with structured knowledge (e.g., knowledge graphs) can significantly improve their performance, especially in inference and reasoning tasks
\cite{feng2023knowledge, sun2023think,pan2024unifying,wang2024knowledge}. Several works use LLMs to parse text into structured representations, manipulate these representations, and (optionally) apply the LLM again to translate the result into text %, especially in inference and reasoning tasks 
\cite{yang2023coupling,zelikman2023parsel,besta2024graph,zhang2025sr}.

In this work, we show that surprisingly, incorporating structure can also improve LLMs' \emph{creativity} and \emph{diversity}.
We stress that we do not mean creativity and diversity on the lexical (token) level; rather, we want the model to be creative on a more abstract level, within the realm of concepts (or the ``landscape of ideas'', so to speak). 

{Our approach is illustrated in Figure~\ref{fig:pipeline}. 
% \dnote{but this paragraph really doesn't explain what's in Fig 1, you're missing the whole generate/evaluate part}\mnote{tried to link better to Fig 1}
Similar to parsing-based approaches,
% \cite{tian2024largel, zhao2023large, li2024simple}, 
we start by deriving structured representations from textual inputs using an LLM. 
% \dnote{why are we citing those after we just cited a whole bunch of methods that do this two paragraphs before?}
% Next, we manipulate these structured representations to reach creative parts of the idea space. 
We then manipulate these structured representations externally, thereby generating new ideas while systematically exploring creative regions of the idea space.}
{Inspired by the human creative process, we focus on \textbf{recombination} -- a fundamental principle in creativity research, which posits that novel ideas often emerge by merging existing concepts in unexpected ways \cite{guilford1967nature, utterback1996mastering, ahuja2001entrepreneurship}.  
For example, combining pizza preparation methods with the flavors of alfajores cookies might yield a brand new  ``alfajores pizza''; combining a sofa with a bookshelf might result in multifunctional furniture. % optimizing both comfort and space efficiency. 
To recombine structured representations of existing ideas, we employ an edit-distance algorithm, and focus on representations midway through its transformation steps. 
We 
% systematically explore 
sample from the space of recombinations,
evaluating candidates for novelty and value \cite{finke1996creative, sawyer2024explaining}. 
% Candidates that score highly 
Those deemed most promising are then translated into natural language.}

% \mnote{Maybe we should say something about that in the evaluation phase we incorporate more domain-specific knowledge?}
% \dnote{why here?} \mnote{agree!}
%\dnote{ We manipulate the structure (recombination) and sample on this space}

\begin{figure*}[h]
    \centering
    \includegraphics[width=\textwidth]{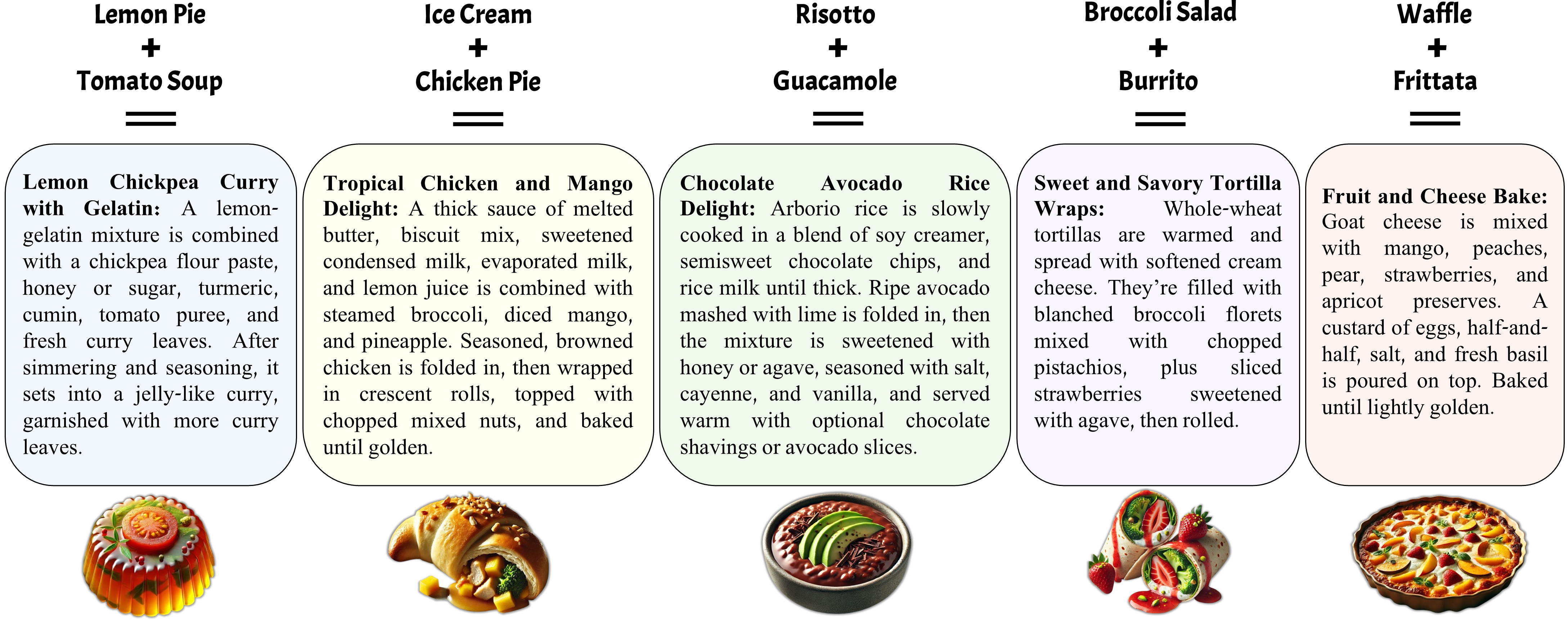}
    \caption{\label{fig:main}
    Examples of new recipe ideas generated by \alg{}. Each example consists of a pair of input dishes and their best corresponding generated recipe idea. The generated recipes are presented as concise summaries to conserve space.  Images were generated using OpenAI’s DALL·E.
    }
\end{figure*}

We demonstrate our paradigm in the culinary domain, introducing \alg{}, a model for creative generation of recipes.
Figure~\ref{fig:main} presents examples of recombinations generated by it. %, showcasing pairs of dish inspirations alongside new recipes created through recombination.
%
%\dnote{I'll write this after I go through the eval section again, but should I say that we lose a bit at value?}
%demonstrating a clear advantage in creative generation. 
%
Beyond the scope of cooking, we believe this paradigm holds promise for extending creative and diverse generation to domains where suitable structured representations and value criteria can be defined (e.g., procedural texts, drug design, music generation; see Section \ref{sec:discussion}).
% a wide range of domains, from product design, narrative construction, and scientific discovery to artistic composition. \mnote{tone down: "...could extend to domains where suitable structured representations and value criteria can be defined (e.g., ...), a point we discuss in Section \ref{sec:discussion}}. 
Our contributions are:

%Beyond introducing fresh, unexpected possibilities into a domain, creative generations could also simulate novel data cases in scenarios where real-world data is scarce, and serve as ``creativity boosters'' for human collaborators.

%suggestions for contributions :)
%I'll go over them again after going through eval. I think I'll go there now, actually

%We introduce a novel paradigm designed to enhance LLM creativity, based on the idea of extracting structured representations from natural-language inputs and manipulating them 

\begin{compactitem}
    \item We introduce a novel paradigm to enhance LLM creativity
    by extracting structured representations, applying cognitive inspired manipulations, and decoding the results back into natural language, thus going beyond superficial token-level variation.
    \item We propose a new recombination operator based on edit distance, which enables controlled blending of structured ideas by partially transforming one representation into another.
    \item We demonstrate our approach in the culinary domain with \alg{}, a model that recombines recipes to generate creative ones. 
    \item We curate a 5K-recipe dataset generated by \alg{}, providing a valuable resource for future work on creative generation. 
    We make both the code and data publicly available.\footnote{\href{https://github.com/moranmiz/Cooking-Up-Creativity}{https://github.com/moranmiz/Cooking-Up-Creativity}\label{fn:link}}
    % \footnote{Code and data are available at [link redacted for anonymity].\label{fn:link}}  
    % We share the code and data.
    % { for reproducibility and further research}
    \item  Through systematic experiments, we show that \alg{}’s generations are significantly more \textbf{diverse} compared to baseline SOTA LLM outputs. Most recipes generated by both models are deemed valuable (appropriate and coherent), although the baseline achieves better scores on an open-ended task. Most importantly, our outputs significantly surpass the baseline in terms of novelty, resulting in more \textbf{creative} culinary ideas. These findings are supported by both automated metrics and domain expert evaluations.
\end{compactitem}

% %Experiments comparing our model's results to those of \cgpt{}
% show our generations are more diverse. Most recipes generated by both models are deemed valuable (i.e., they make sense), with \cgpt{} achieving better scores on an open-ended task. More importantly, our outputs significantly surpasses \cgpt{} in terms of novelty, resulting in more creative dishes. 

% \begin{compactitem}
%     \item We introduce a novel paradigm designed to enhance LLM creativity, based on the idea of extracting structured representations from natural-language inputs and manipulating them 
%     \item We demonstrate our paradigm in the culinary domain. While designed for recipes, we believe this paradigm could be generalized to other creative domains.
%     \item Our model generates a 5K-recipe dataset.  We make our code and data publicly available.\footnote{The link is redacted to preserve anonymity.}
%     \item Through systematic experiments, we show that our system’s outputs are more diverse, less repetitive, and significantly more surprising -- yet mostly appropriate -- than those produced by a baseline state-of-the-art LLM (\cgpt{}).
% \end{compactitem}

\section{Background: Human Creativity} 
\label{sec:background_and_definitions}
% \begin{quotation}
% \noindent ``Any creative act is a leap when the mind connects two unrelated ideas.'' \\
% \textit{--\citet{koestler1964act}}
% \end{quotation}
The field of human creativity has been extensively studied, identifying numerous principles that drive innovation. In designing our model, we relied on the following principles:

\xhdr{Generation \& Evaluation}
%Human creativity has been widely studied in cognitive science and psychology, revealing that 
%Creative thinking often unfolds in distinct mental stages \cite{wallas1926art, bransford1993ideal}. 
A common yet effective model of creative thinking is the two-stage process {\textit{generation \& evaluation}}, which suggests that creativity begins with divergent thinking (free idea generation), followed by convergent thinking, where the most promising ideas are selected and refined \cite{finke1996creative, sawyer2024explaining}. We incorporate this as the conceptual backbone of our model, implementing a generative component that produces a broad set of  ideas, followed by an evaluative component that identifies those with the greatest creative potential.

\xhdr{Recombination of Ideas}
We base our work on a prominent idea-generation method: \textit{recombination}, where elements from existing ideas are  merged to create novel concepts \cite{koestler1964act,guilford1967nature}. Our model strategically recombines elements from pairs of existing seed ideas to spark unexpected connections.

\xhdr{Creativity Assessment: Novelty \& Value}
After generating many ideas, the challenge is determining which are genuinely creative. %However, measuring creativity is difficult. 
Numerous studies have examined the complexities of assessing creativity in both humans and computational systems \cite{said2017approaches, lamb2018evaluating}.
A widely accepted definition of creativity frames it as the intersection of \textit{novelty} and \textit{value} \cite{mumford2003have, boden2004creative, boden2009computer, lamb2018evaluating}. 
Novelty ensures that an idea is surprising or unconventional, while value signifies it is useful in its intended context. 
% In the culinary domain, a recipe may be considered creative if it introduces unexpected ingredients or techniques (novelty) while still resulting in a delicious and practical dish (value). %Based on this framework, we assess newly generated recipes along two dimensions: (1) Is it sufficiently distinct from known ideas? (2) Does it meaningfully fulfill its intended purpose?

\xhdr{Measuring Novelty and Value Automatically.}
Novelty can be assessed by identifying how uncommon an idea is within a dataset \cite{heinen2018semantic, kenett2019can, doboli2020cognitive}. 
% One common strategy explored in prior work used embedding-based outlier detection to highlight ideas that deviate from established norms \cite{dunbar2009creativity, harbinson2014automated}. 
Evaluating value, however, is  highly domain-dependent, often considered the ``holy grail'' of computational creativity \cite{boden2004creative, ritchie2007some, jordanous2012evaluating}. Thus, we consider value assessment as a domain-specific task.

\section{Problem Definition}
\label{sec:problem_definition}
Innovation often involves combining existing ideas to create novel ones. This process, often referred to as ``conceptual blending'' or ``creative recombination,'' is central to innovation, and the focus of our work. We now introduce {\bf key elements} of our formulation.

%In our approach, we create novel recipes by recombining existing seed recipes. More formally,
Given a domain where ideas can be expressed in a structured format (e.g., cooking recipes, instruction manuals, computer programs),
% , music compositions)
let $\mathcal{I}$ denote the theoretical set of all possible ideas within that domain -- both existing and yet-to-be-discovered. $\mathcal{I}$ represents the entire conceptual space of ideas that adhere to the domain's structural and logical constraints, encompassing all valid possibilities. In addition, let $I\subset\mathcal{I}$ be a set of ideas that have been recorded
% , recognized, 
or are known within the field. %documented ideas within the domain. While $D$ is not necessarily exhaustive, it represents the set of 
% \begin{displaymath}
% \mathcal{D} =  \{ d \mid d \text{ is a valid idea in the domain} \}
% \end{displaymath}

\remove{
A non-valid idea in the cooking domain might be a recipe with illogical steps (e.g., not cooking raw chicken) or an inedible result (e.g., soup thickened with sand). 
% In music composition, it could be a sequence of notes that breaks fundamental melodic or harmonic principles, and in programming, it could be code that fails to run.
In furniture design, it could be a design that disregards basic principles of stability or usability, and in storytelling, it could be an incoherent plot that contradicts its own internal logic.}

\begin{definition}[\bf Recombination Function]
    Recombination function $\mathcal{C}$ takes as input two structured ideas $i_a, i_b \in I$ and produces a set of new combinations $I_{ab} \subseteq \mathcal{I}$ such that each $i\in I_{ab}$ is a different mixture of $i_a, i_b$. 
    %$D_{AB}=\mathcal{C}(i_A,i_B)\subseteq \mathcal{D}$ such that each idea in $D_{AB}$ 
\end{definition}

The exact definition of ``mixture'' depends on the representation. For example, when we transition from representation $i_a$ to representation $i_b$ with a minimal edit distance procedure, the intermediate steps can be viewed as mixtures of $i_a$ and $i_b$, blending elements of both in varying proportions as we move through the transformation. 

\remove{
For example, in the cooking domain, a common structured format used to express ideas is a tree. 
}

\begin{definition}[\bf Evaluation Function]
    The output of a recombination is a set of potential innovations $I_{ab}$, which can be evaluated with evaluation function $E: \mathcal{I} \rightarrow \mathbb{R}$.  
 The innovation can be evaluated based on criteria such as novelty and utility. 
\end{definition}

\begin{definition}[\bf Retrieval of Ideas from Descriptions]
Ideas are often expressed in different levels of abstraction and granularity. Let $m$ be a function that matches an idea description $d$ to relevant known ideas from $I$, $m(d) \subseteq I$. For example, $m$ could match the textual description ``lasagna'' to all lasagna recipes.   
\end{definition}

\noindent The formal optimization problem can thus be stated as:
Given two idea descriptions $d_a, d_b$, find
\begin{equation}
    \argmax_{i\in \mathcal{C}(i_a, i_b) \ \mid \ i_a\in m(d_a),\; i_b \in m(d_b)} E(i) \nonumber
\end{equation}

% \mnote{Does the equation in the following format actually look better?}
% \noindent The formal optimization problem can thus be stated as:
% Given two idea descriptions $d_a, d_b$, find
% \[
% \begin{aligned}
% &\arg\max_{i} \quad E(i) \\
% &\text{subject to} \quad i \in \mathcal{C}(i_a, i_b), \\
% &\phantom{\text{subject to} \quad} i_a \in m(d_a),\; i_b \in m(d_b)
% \end{aligned}
% \]

Figure~\ref{fig:main} illustrates examples of generated ideas in the domain of cooking recipes, along with the idea descriptions used to create them. For example, combining broccoli salad recipes and burrito recipes resulted in a recipe for a tortilla filled with cheese, broccoli, strawberries and pistachios.

%\begin{equation}
%    \max_{\substack{d\in \mathcal{C}(d_a, d_b) \\ d_a\in m(i_a), d_b \in m(i,b)}} E(d) 
%\end{equation}

\remove{
\begin{definition}[\textbf{Inspiration Set}] 
We define an \textbf{inspiration set} $I$ as a small, finite, curated subset of $D$ that represents specific themes, styles, or patterns of interest. This set provides a focused source of inspiration for generating new ideas.
\end{definition}}

\remove{
\begin{definition}[\textbf{The Idea Generation Task}]
Given one or more inspiration sets $I_1,...,I_n$ drawn from $D$, our goal is to generate a new idea set $G$ such that for every $g\in G$: (1) $g\in \mathcal{D}$, i.e., $g$ is a valid idea in the domain; 
% (2) $g\notin D$, i.e., $g$ is a novel idea not yet discovered in the domain; 
(2) $g$ is unlikely to be documented in $D$;
% (3) $g$ is influnced by every one of the inspiration sets $I_1,...,I_n$.
(3) $g$ incorporates elements from each inspiration set $I_i$, ensuring it meaningfully reflects all provided inspirations. \gabis{Aren't we combining two ideas $i_1, i_2 \in I$? Why do we need to work with two sets as inputs?} \mnote{two sets - using them we make a lot of combinations from which we choose only the best to translate back}
\end{definition} 
\cnote{I don't like def 2, not sure why (2) is true and everything is very confusing. Also, I agree with Gabi's comment.}
}

\section{Model}
\label{sec:algorithm}
%\gabis{This is a very complex model, and I worry that it may seem arbitrary (why did we choose this specific architecture?). I think we can strengthen the connection to the principles we described in the previous section, why exactly do they entail this specific architecture?}

In this section, we introduce {\alg{}}, 
% \gabis{I think this was stylized differently in the intro},
% (\textit{\textbf{Creative REcipe Merging Engine}}),
our model for automatically generating innovative recipes.\footref{fn:link} Figure~\ref{fig:pipeline} illustrates our methodology. The input consists of two seed inspirations (idea descriptions $d_a$, $d_b$).\footnote{Note that more inspirations can be used if desired.}
Each idea description is mapped to a set of specific recipes (instances of the idea, for example different lasagna recipes; 
% see Section 
\S~\ref{algo:sampling}). 
% $A$ and $B$, 
%each containing multiple instances for a specific dish (e.g., lasagnas and chocolate pies; ). 
% \cnote{A and B were not mentioned in the problem definition, please add them to link between the two sections} 

These recipes are first translated into tree representations using an LLM (step (I) in Figure \ref{fig:pipeline}, \S~\ref{algo:text_to_tree}). 
% To generate new ideas, w
We recombine these trees to produce new candidate ideas with a minimal edit-distance algorithm (step (II), \S~\ref{algo:edit_distance}). Then, 
% utilizing the Novelty \& Value principle, 
% we correct the candidate ideas, filter out nonviable ones and rank the remaining 
we refine the candidate ideas to assess their value and rank them based on their novelty scores (step (III), \S~\ref{algo:eval}). 
%
% Finally, the highest-ranked trees are translated back into natural language recipes and refined using an LLM. 
Finally, the highest-ranked trees are translated back into natural language recipes using an LLM,
which leverages its commonsense and world knowledge to fill in missing details and produce coherent recipes (step (IV), \S~\ref{algo:tree_to_text}). We now provide more details about each step.
% \begin{enumerate}
%     \item \textbf{\textit{Seed Ideas as Input}}: We begin with two sets of “inspiration” ideas (note that more sets can be used if desired). These seed ideas form the starting point for generating new concepts.
%     \item \textbf{\textit{Translation to Mathematical Representation}}: Seed ideas are converted into a structured mathematical form using an LLM. While our focus is on tree-based representations, other formal structures could be used.
%     \item \textbf{\textit{Generation}}: to generate new ideas we algorithmically recombine these representations to produce new candidate ideas.
%     \item \textbf{\textit{Evaluation}}: Using the Novelty \& Value principle, we refine, filter, and rank ideas based on the statistical rarity of their elements in the domain repository, prioritizing surprising yet meaningful ones.
%     \item \textbf{\textit{Translation to Natural Language and Refinement}}: The highest-ranked representations are translated back into natural language using an LLM. This step includes refining, correcting, and enriching the ideas with necessary details.
% \end{enumerate}

\subsection{Sampling Seed Ideas (Step I)}
\label{algo:sampling}

% We base our work on Recipe1M+, which consists of more than one million recipes available on the Internet.

We selected the 100 most popular dishes (e.g., chicken salad, cheesecake) that span different categories (e.g., appetizers, desserts, main courses) from the Recipe1M+ dataset \cite{marin2021recipe1m+}. 
On average, each selected dish is associated with 2,576.33 recipes in the dataset.% (std: 2,580.819; min: 870 for lemon pie; max: 17,004 for pasta) \gabis{TMI in my opinion, I think we remove the parenthesis}. 

To keep the financial costs of using an LLM manageable in the next stage of our model, we sampled 30 recipes per dish, resulting in a total of 3K recipe samples.
To ensure both diversity and representativeness, we selected 15 recipes at random to capture the typical version of each dish, and 15 more to maximize diversity. 
Implementation of diversification can depend on the representation and the domain; in our case, we use the GMM algorithm over recipe embeddings obtained from a Sentence-BERT model fine-tuned on recipe data (see details in Appendix~\ref{algo:sbert_finetuned}), which identifies the dish's embedding centroid and iteratively selects recipes farthest from both the centroid and previously chosen samples \cite{ravi1994heuristic}.

%Specifically, to enhance diversity, we compute recipe embeddings using a fine-tuned Sentence-BERT model trained on recipe data. We then identified the centroid of the dish’s embeddings and iteratively selected recipes farthest from both the centroid and the previously chosen recipes. This approach follows a well-known 2-approximation greedy heuristic for MAX-MIN sampling, commonly referred to as the GMM algorithm \cite{ravi1994heuristic}.

\begin{figure}[t!]
\includegraphics[width=\linewidth]{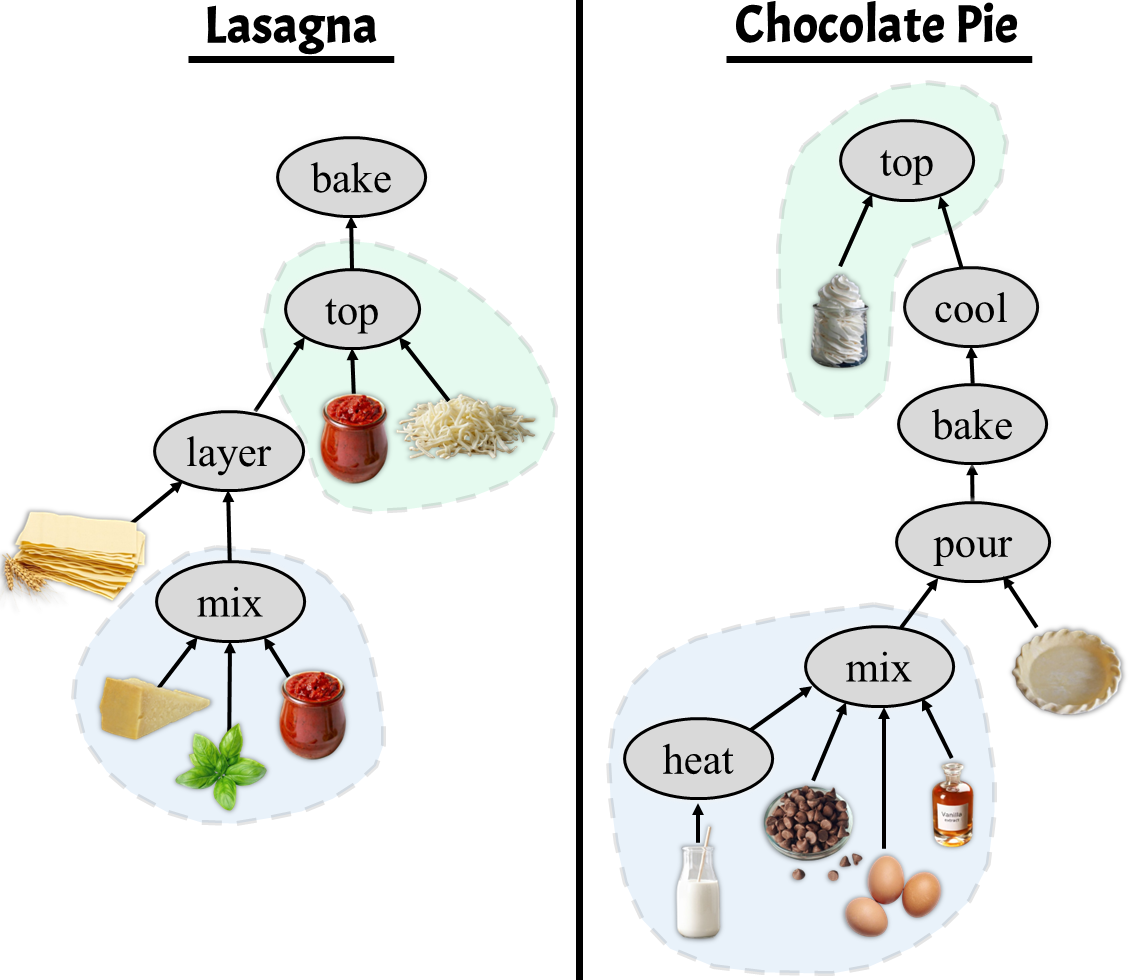}
\caption{\label{fig:recipe_trees}
Tree representations of lasagna and chocolate pie recipes. Analogous parts are highlighted, showing structural similarities that the minimal edit distance algorithm is more likely to preserve when transforming one tree into another.
}
\end{figure}

\subsection{Text to Tree (Step II)}
\label{algo:text_to_tree}

Cooking recipes, like experiments, assembly manuals, and game instructions, are procedural texts. These texts typically consist of a sequence of steps accompanied by the objects needed to perform them. A common way to represent procedural texts is as a tree \cite{jermsurawong2015predicting, maeta2015framework}, where leaf nodes correspond to needed objects (in our case, ingredients), and  internal nodes represent the actions performed on them. Figure~\ref{fig:recipe_trees} shows simple lasagna and chocolate pie recipes represented as trees.

%A tree structure is convenient as it helps capture dependencies between steps and makes it simpler to manipulate and combine procedures. 
% To translate a recipe text into a tree, 
To parse recipe text into tree,
we prompted \cgpt{} %, leveraging its extensive world knowledge and code generation capabilities 
with a chain-of-thought approach. See Appendix~\ref{app:text_to_tree_details} \&~\ref{app:text_to_tree_prompts} for full details and corresponding prompts. The total cost of generating tree representations for 3K recipes was approximately \$40.

An initial pass revealed that 1,347 (44.9\%) of the resulting trees were invalid due to issues such as orphan nodes, multiple outgoing edges from a single node, or incorrect edge directions. 
To address this, we implemented a correction step where we removed problematic edges and instructed the model to reconsider them. This improved validity to 95\% (2,850 trees).
We then evaluated the final trees using 50 random recipes. A gold-standard tree was created for each recipe, and we automatically compared the predicted trees with the gold trees in terms of node and edge matching. For nodes, we achieved 0.985 precision, 0.956 recall, and F1-Score of 0.969. For edges, we obtained 0.951 precision, 0.909 recall, and  F1-Score of 0.93. 
Overall, these results demonstrate the effectiveness of our approach in translating procedural texts into a structured tree representation, making them suitable for further manipulation and analysis.

\begin{figure*}[h]
    \centering
    \includegraphics[width=\textwidth]{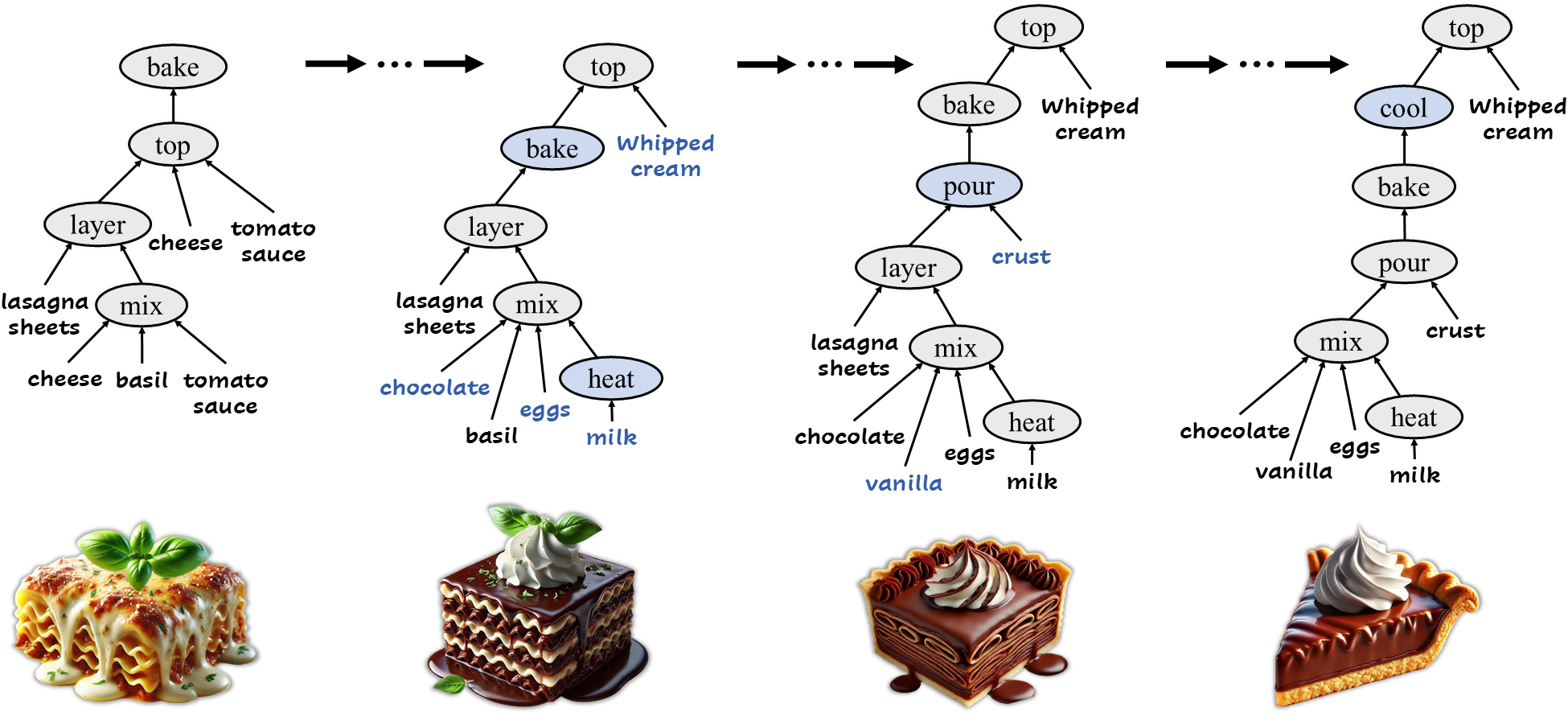}
    \caption{
    Illustration of our tree-based edit-distance approach for generating new recipe ideas by transforming a simple lasagna tree (left) into a simple chocolate pie tree (right). Intermediate “merged” trees are produced for each edit step, yielding novel dishes such as a \textit{basil chocolate lasagna} or a \textit{chocolate lasagna encased in a pie crust}. Recipe images were generated using OpenAI’s DALL·E model.
    }
    \label{fig:edit_distance_process}
\end{figure*}

\subsection{Generate Ideas (Step III)}
\label{algo:edit_distance}

In this section we operationalize the recombinaction function $\mathcal{C}$.
We generate novel recipes by blending recipe trees with a minimal edit-distance algorithm. 
%To illustrate our approach, we first present a string-based minimal edit distance example. Given two strings $a$ and $b$ over an alphabet $\Sigma$, their minimal edit distance $d(a,b)$ is the smallest number of insertions, deletions, and substitutions required to transform $a$ into $b$. For instance, the minimal edit distance between the words “SPOON” and “FORK” is four, achieved by substituting letters and deleting one character. Notably, tracing the transformation steps,  SPOON → SPOOK → SPORK → SFORK → FORK, reveals that stopping midway in this sequence (e.g., at “SPORK”) produces a valid new word that blends elements from both “SPOON” and “FORK”. This example underscores the central idea of our method: 
The key idea behind this method is that by examining the step-by-step transformation between two concepts, we can discover intermediate forms that blend features of both.

In the case of recipe generation, we employ the Zhang–Shasha algorithm, which computes the minimal edit distance between trees \cite{zhang1989simple, bille2005survey}. Given two recipe trees $i_a$ and $i_b$, we compute their minimal edit distance and record all operations required to transform $i_a$ into $i_b$. Each operation sequence produces intermediate trees that represent novel ``merged” ideas, from which we randomly select one as our new recipe. Figure~\ref{fig:edit_distance_process} illustrates an example sequence transforming a simple lasagna tree into a simple chocolate pie tree. One intermediate variant might be a chocolate lasagna with basil; another could feature a chocolate lasagna encased within a crust. 

A key advantage of the minimal edit distance approach is its ability to preserve the structural roles of ingredients and cooking steps. For example, in Figure~\ref{fig:recipe_trees}, both the lasagna and chocolate pie recipes include a “topping” action (marked in color). Using a minimal edit distance 
% algorithm 
ensures that inserting whipped cream aside of tomato sauce and cheese is more likely, as placing whipped cream elsewhere would increase the overall edit cost. See 
% full 
implementation details in Appendix~\ref{app:tree_edit_distance}.

Note that stopping at different points in the transformation process can create unique dishes (see Figure~\ref{fig:edit_distance_process}). Additionally, shuffling the order of edits can generate entirely new intermediate ideas.

\subsection{Evaluate Ideas (Step IV)}
\label{algo:eval}
%\gabis{This subsection is unclear to me. why are we talking about filtering ideas and ranking them? Is this all done on top of the predicted new recipes that our model outputs? I think we need to clarify what's the relation between what we talk about here and the previous one. After reading on a little bit, I understood it for myself (I think). I think that it would be very helpful to point to Figure 2 in every subsection here to anchor the reading. Also, use the exact terms we see in that figure. I don't think that ``Evaluation'' is a good name, it threw me off, thinking we're now talking about evaluating the output of the whole model. In the figure we say ``Correct'', ``Filter'', ``Rank'', I think we should use the same terminology (although ``correct' isn't clear to me).}

%\dnote{tie to new objective}
Now that we have a set of candidate innovations created through recombination, we evaluate each generated recipe. As noted in Section \ref{sec:problem_definition}, the evaluation function $E$ should take into account novelty and value. Specifically, we chose to view value as a constraint, and novelty as the optimization objective; i.e., we wish to rank by novelty all candidates that pass a value threshold (i.e., make sense). 

%Implementation, as before, is somewhat domain-dependent. 
In the culinary domain, a recipe is considered creative if it introduces unexpected ingredient or technique combinations (novelty) while resulting in a delicious, coherent dish (value, utility). We next operationalize these criteria. Our implementation is specific to recipes, but we believe the principle can be generalized to other domains.

%we propose a three-step pipeline: (1) \textbf{improve} the ideas for taste, %(coherence is addressed later by the LLM), 
%(2) \textbf{filter} out suboptimal or trivial variations of known recipes, and (3) \textbf{rank} the remaining ideas based on the statistical rarity (surprise) of their elements.
%\dnote{explain which step is value and which is novelty}

\subsubsection{Value Constraint} %\gabis{I don't like this formatting which is non-conventional, why not as a paragraph?}
%\dnote{if you agree, rename in the figure}
%The trees we obtain from the recombination function $\mathcal{C}$  sometimes display unlikely combinations of ingredients. In this section, we perform a simple greedy heuristic to ameliorate this issue. This step is optional. 

To assess value (taste) of a recipe, we follow \citet{varshney2019big}, and check if its ingredients pair well with each other. 
{We use {flavorDB} dataset \cite{garg2018flavordb}, which catalogs taste molecules for raw ingredients.}
According to {this work,} two ingredients pair well if they share a larger proportion of taste molecules; we compute a Jaccard-based pairing score for each pair of {raw} ingredients. 
{Since this dataset does not cover processed ingredients, we also use {FoodData Central} \cite{fukagawa2022usda} to infer their raw components. We define the pairing score between two composite ingredients as the lowest score among their constituent raw-ingredient pairs. After exploration, we chose 0.3 as our value threshold.}
If the score falls below {this} threshold, we consider the pairing problematic. We iteratively remove the ingredient with the highest number of low-scoring pairings until no further collisions remain (alternatively, one might choose to remove these candidates altogether). 
See details in Appendix \ref{app:conflict}.

%\subsubsection{Filtering Idea Representations}
%After ensuring that each recipe is free of major taste conflicts, we filter out suboptimal or trivial ideas. 

\subsubsection{Ranking by Novelty}

Our intuition is that novel recipes include ingredients and actions that do not often appear together. Thus, we formulate a measure of surprise, inspired by 
% the principles of 
the inverse document frequency (\textbf{idf}) concept in \textbf{tf-idf} \cite{ramos2003using}.

%To rank the remaining recipes, we measure how novel each idea is relative to our domain repository. We define a ``surprise metric’’ that parallels 
%
Let $T$ be a recipe tree with nodes $E_T$, where each node represents an \textbf{element} (either an \textbf{ingredient} or a \textbf{cooking action}). Let $N_e$ be the number of recipes in the repository that include element $e$. For each $e' \in E_T\setminus \{e\}$, let $df_e(e')$ denote the number of recipes containing both $e, e'$. We  define:\footnote{To prevent score inflation due to typos or extremely rare ingredients, we exclude any ingredient that appears exceptionally infrequently in the entire dataset.}
\begin{displaymath}
idf_e(e’) = log\Bigl(\frac{N_e}{df_e(e')}\Bigr)
\end{displaymath}
A higher $idf_e(e’)$ score indicates that $e'$ is more unique relative to recipes containing $e$.

To compute novelty of an element $e$ in $T$, we compute its $idf_e$ score with respect to each of the other elements in $T$, and sum the top 10 values. To compute the overall novelty of $T$, we use the sum of the top-10 element-level novelty scores. A chicken and mango delight recipe (Figure \ref{fig:main}) was deemed novel due to ingredient combinations like mango, crescent roll, nuts and broccoli, and actions such as steaming, blanching and unrolling.

%We compute the \textbf{element novelty} $Novelty_e(T)$ by summing the top ten $idf_e$ scores for elements in $T$. The overall \textbf{novelty score} of a recipe is then the sum of these top-ten element scores across all elements. In this way, we reward recipe ideas composed of elements that are collectively rare within the repository. 

%\dnote{I didn't understand this last paragraph at all}
%\mdel{Importantly, we designed this metric to be comparable across different dish pairs, allowing us to extend our generator beyond individual fusions. By applying the process to many specific pairs, we can generalize it to propose novel recipes independent of any particular pair (see Section~\ref{sec:evaluations}).}

%\madd{Note that this novelty score definition considers only the tree elements (ingredients and actions) and does not depend on the specific pair of dishes being recombined. This allows for comparing the novelty of recombinations derived from different dish pairs.} \mnote{is it clearer now?}

\subsection{Tree to Text (Step V)}
\label{algo:tree_to_text}

After identifying the highest-ranked trees, we convert them back into natural-language recipes using an LLM. Similar to Section~\ref{algo:text_to_tree}, we employ a chain-of-thought approach, instructing \cgpt{} to 
% first 
translate the tree (encoded in DOT format) into a structured recipe, including a title, ingredient list, and step-by-step instructions. We then prompt the model to refine and correct the text for coherence, fluency, and consistency (see Appendix~\ref{app:tree_to_text_prompts} for full prompts). 
Processing 1,000 recipe trees in this step required a total cost of approximately \$42.

We note that the LLM plays a key role in \emph{surface realization}. Specifically, it draws on commonsense and domain knowledge to (1) \textbf{fill in missing details} (e.g., suggesting plausible ingredient quantities or cooking times), and (2) \textbf{correct inconsistencies or omissions} introduced during recombination (e.g., restoring a step for cooking raw chicken if it was omitted).

\section{Creative Recipe Dataset}
\label{sec:dataset}
%\mnote{ADDED THIS SECTION}
% \paragraph{ Space of Possibilities.} \gabis{try to have a proper sentence here, what about the space of possibilites? do we aim to filter them?}
%Our approach allows for a wide range of new recipe ideas. As shown in Figure~\ref{fig:edit_distance_process}, 
%For instance, adding a crust to the lasagna early in the transformation process may result in a dish that tastes like lasagna but has the structure of a pie. Extending this method to three or more recipe trees can further expand the range of possible recombinations, making this tree-based edit-distance approach a powerful tool for automated idea generation.

% To explore this space, we constructed an intermediate dataset of merged trees. As mentioned earlier, we sampled 30 recipes for each of the 100 most common dishes and converted them into tree structures. We then paired $\sim$1,000 dishes from different categories. For each dish pair, and each recipe pair within it (one from dish A and one from dish B), we generated three merged trees in both directions (A → B and B → A), shuffling the order of edits each time. This process resulted in up to 5,600 merged trees per pair and $\sim$5.5 million in total.

% For each of the $\sim$1,000 dish pairs for which we generated new ideas (as described in Section~\ref{algo:edit_distance}), we selected the five highest-ranked trees and converted them into recipes, forming our dataset of newly generated recipe ideas.
% This staged approach required a total cost of approximately \$42 to process 1000 recipe idea trees.

We used our model to generate a new dataset of recipes. We first identified 100 popular dishes spanning different categories. For each dish, we sampled 30 recipes and converted them into tree structures. Next, we sampled 1,000 dish pairs, ensuring that pairs include dishes from different categories. For each pair of dishes, $d_a$ and $d_b$, every recipe pair in $m(d_a)\times m(d_b)$ was used to generate six blended trees, producing up to 5,600 trees per dish pair ($\sim$5.5M in total). We selected the five highest-ranked trees from each set and converted them back into natural language recipes, resulting in a dataset of 5K  recipes.

\section{Evaluation}
\label{sec:evaluations}

We now turn to evaluate \alg{} by investigating the following research question: 
\textbf{\textit{How do the recipes \algbold{} generated compare to those generated by a SOTA LLM (\gpt{})?}}

To answer this, we examine how  our outputs compare to those of \gpt{} in two key aspects: \textbf{diversity}  and \textbf{creativity}. For diversity, we explore whether our approach mitigates the well-documented issue of repetitiveness in LLMs. %(which is a critical challenge in creative generation). Second, we assess how our outputs compare to those of \gpt{} in terms of  \textbf{creativity} -- 
For creativity, we look for outputs that are both valuable (make sense) and novel (unexpected). 
We compare our model's outputs with those of GPT-4o\footnote{Version: gpt-4o-2024-08-06 (latest stable version of the model at the time of writing this paper).} on two tasks: 

\noindent{\bf Experiment 1.}
We evaluated \gpt{} and \alg{} on their ability to generate creative recipes \textbf{combining a given dish pair}. The evaluation included 10 randomly selected dish pairs. For each pair, we selected 5 recipes generated by \gpt{} 
and compared them to the top 5 recipes generated by \alg{} (50 recipes per model).
\\ [0.2cm]
\noindent{\bf Experiment 2.}
To broaden the scope of our analysis, we wish to evaluate the most creative recipes \gpt{} and \alg{} could generate in general, \textbf{without limiting them to a given input pair}. We used 100 recipes from each model. 
% We instructed \cgpt{} to generate 100 different creative recipes in a single chat session.
For \alg{}, we selected 100 recipes from our 5K-recipe dataset, which includes a novelty score for each generated recipe. To ensure that the dishes do not come from a very small number of inputs, we employed simulated annealing \cite{bertsimas1993simulated}, maximizing recipe novelty while enforcing a constraint about the maximum number of appearances of each dish.

Note that Experiment 1 is a more unconventional task, one that the model is less likely to have encountered during training. We evaluate the outputs of both experiments using  qualitative and automated analyses as well as human annotations.

\subsection{Experimental Details}
\label{eval:experimental_details}

Here we describe the process of generating recipes for the \gpt{} baseline as well as the human evaluation setup.

\xhdr{Recipe Generation for the \gpt{} Baseline}
LLMs are known to be sensitive to prompt paraphrases \cite{Sclar2023QuantifyingLM, mizrahi2024state, voronov2024mind}. 
To ensure a fair and competitive baseline, we conducted a thorough prompt design process.
First, we constructed large, diverse pools of prompt paraphrases tailored for each experimental task. Specifically, we developed 104 prompts for Experiment 1 and 114 prompts for Experiment 2. These prompts varied systematically along several axes: structure (explicit steps vs. open-ended, explicit chef role vs. no role, etc.), length (concise vs. detailed instructions), creativity-oriented language, and  creativity-related constraints.\footnote{All prompt variations are  publicly available alongside the dataset and code for full reproducibility.}
% Over 85\% explicitly guided GPT-4o toward novelty (e.g., by requesting novel ingredient combinations or innovative cooking methods). Over 75\% explicitly encouraged valuable outputs, either directly (e.g., emphasizing harmonious dishes, flavor balance, ingredient compatibility, and clarity of instructions) or indirectly (by assigning GPT-4o the role of a culinary expert). 

We tested different temperature settings (ranging from 0.0 to 2.0) for a sample of these prompts. For $t>1$, \gpt{} often produced gibberish. In line with the findings of \citet{peeperkorn2024temperature}, we observed  improvements in novelty at higher temperatures, although they were rather subtle.  We therefore selected $t=1$ 
% \dnote{reaching this decision how?}
as the highest temperature that produced coherent recipes. %, noting that higher values resulted in gibberish. 

To assess the diverse prompt set, we conducted a small evaluation.   For Experiment 1, we generated a recipe for each of three sampled dish pairs and every prompt (resulting in 312 recipes). For Experiment 2, we generated three recipes independently per prompt (resulting in 342 recipes).
%Next, we carefully analyzed the initial \gpt{} outputs from these diverse prompt sets. 

Qualitative analysis of the generated recipes revealed significant repetition: For experiment 1, \gpt{} consistently generated recipes that prepared each component separately and combined them at the end of the process in the same superficial manner. For example, all burger-waffle recipes involved cooking the burgers and waffles independently, then simply stacking the burger between two waffles. This observation is supported by high average Self-BLEU scores (BLEU-2=0.929, BLEU-3=0.877, BLEU-4=0.818), and high cosine similarity between recipes of the same pair (average $\sim$0.90 across sampled dish pairs).

For experiment 2, \gpt{} frequently reused known culinary dish concepts with minor variations (e.g., marinated proteins served with quinoa/rice in approximately 60 recipes). This observation is again supported by high self-BLEU scores (BLEU-2=0.912, BLEU-3=0.829, BLEU-4=0.739), and high similarity to the large corpus of existing recipes, with average cosine similarity to nearest corpus neighbor of 0.853 (std=0.035).

Due to this pronounced repetition, using multiple prompts in the experiments risked redundant outputs, weakening the baseline. Therefore, two culinary experts reviewed the \gpt{} outputs to select the \textbf{single best-performing prompt} per experiment -- the one that consistently produced the most creative and varied recipes. Importantly, this means that {\bf the \gpt{} baseline was explicitly  optimized for the experiments' evaluation criterion} (creativity as judged by human experts), 
%aligning it directly with the experiments' evaluation criterion (human judgments), 
 thus giving it a potential advantage over \alg{}.
 
Finally, we generated the baseline outputs for the experiments using these best-performing prompts within a {single chat session}, instructing \gpt{} to produce recipes {differing} from prior outputs to encourage diversity. See the chosen prompts for both experiments in Appendix~\ref{app:exp_prompts}.

\xhdr{Human Evaluation Setup} We used Prolific to recruit and manage a total of 48 participants for both experiments. 
We pre-screened participants based on their cooking experience and frequency, comfort with adjusting recipes, and ability to judge creative outcomes. 
Participants were paid an estimated hourly rate of \pounds9, adhering to ethical compensation standards. 
The participants rated randomly selected recipes drawn from the two models on various aspects related to novelty and value (see the full list of questions in Appendix~\ref{app:human_exp_questions}).
To reduce cognitive load, participants initially viewed concise recipe summaries (also generated by \cgpt{}), with the option to examine complete recipes if desired.
Each recipe was rated by five annotators. Final scores were determined using the median of annotators’ ratings, a standard approach for ordinal data to handle outliers. Novelty and value scores were calculated as the mean ratings of their respective questions; the value score was additionally binarized using a threshold of 4 (on a 1–5 scale) across three related questions. On average, participants rated 31.25 recipes (std=33.827). Participants who consistently completed ratings too quickly ($<$45 seconds per recipe) were excluded to ensure data quality.
% On average, each participant rated 31.25 recipes (std=33.827).
% To ensure quality, we excluded participants who, on average, completed reading a recipe and rating its corresponding eight questions in under 45 seconds.

\xhdr{Human Evaluation Reliability} To confirm the robustness of our human evaluation results, given the inherent subjectivity involved in judging creativity, we computed two metrics commonly used in subjective annotation tasks: Interquartile Range (IQR), capturing numeric consistency among annotators, and Krippendorff’s alpha, assessing overall agreement reliability.
\begin{compactitem}
    \item \textbf{Interquartile Range (IQR):} The average IQR across all questions and recipes was approximately 1.0 (Exp.~1: mean=0.992, std=0.33; Exp.~2: mean=1.008, std=0.372). About 79\% of items in both experiments had an IQR$\leq$1, and only $\sim$3\% of items had an IQR$>$2, indicating generally strong numerical consistency. Analyzing questions individually, value-related questions consistently exhibited lower IQRs ($\sim$0.8–0.9), indicating clearer annotator agreement. Novelty-related questions had slightly higher but still acceptable IQRs ($\sim$1.1–1.3), within the expected range for subjective assessments \cite{margherita2021managing, palomo2025developing}. This indicated annotators interpreted our guidelines consistently, with typical individual variation for subjective judgments.
    \item \textbf{Krippendorff’s alpha:} To further assess inter-rater reliability for ordinal data, we computed Krippendorff’s alpha, obtaining values of 0.201 (Experiment 1) and 0.209 (Experiment 2). While modest, these alpha values are consistent with expectations for subjective tasks like creativity evaluation, where individual interpretations naturally differ, even when numerical responses show overall consistency.
\end{compactitem}

\subsection{Diversity in Generated Recipes}

Diversity plays a pivotal role in creative generation, reflecting how broadly and flexibly new recipes adapt and transform original concepts. 
In this part of our evaluation, we use both qualitative and automated analyses to compare \alg{}’s outputs with those generated by \gpt{}, examining how each model integrates concepts from different dishes and their level of repetitiveness.

\paragraph{Deep vs. Shallow Merging of Recipe Concepts.}
Both \alg{} and \gpt{} demonstrated the ability to merge ideas from different dishes, but we noticed their outputs exhibited different types of integration. \alg{} consistently produced more cohesive integrations, where ingredients from both source dishes were woven together into a single, unified recipe. In contrast, \gpt{} tended toward ``shallow merges'', typically preparing each dish separately before combining them at the end. For example, when asked to create a hybrid of muffins and orange salad, \gpt{} invariably proposed baking muffins, preparing an orange salad, and then serving them together, placing the salad on top or on the side. In contrast, \alg{} generated recipes like a mixed vegetable bake incorporating elements from both dishes, a citrus cake with an orange-based sauce, and a salad that integrated muffin ingredients.

This difference was also reflected in the average ingredient and word counts: \gpt{}’s recipes were nearly twice as long and complex as \alg{}’s. In Exp.~1, \alg{} had on average 12.3 ingredients  (std=3.71) and 240.28 words (std=47.98), compared to \gpt{}’s 24.98 ingredients (std=5.2) and 465.84 words (std=64.8). In Exp.~2, \alg{} used 13.32 ingredients (std=4.39) and 264.15 words (std=51.38) on average, compared to \gpt{}'s 23.79 ingredients (std=3.25) and 371.7 words (std=42.92). For reference, recipes in the Recipe1M+ corpus averaged only 9.33 ingredients (std=4.31) and 168.53 words (std=104.79). This  suggests that \gpt{} tends to generate separate sub-dishes, each with its own ingredients and instructions, and then combine them into a final dish rather than blending the culinary ideas.

\paragraph{Fixation on Structures and Ingredients in \gpt{}’s Outputs.}
\gpt{} showed a strong fixation on certain structures and ingredients. When asked to merge two dishes, it often followed the same approach in multiple attempts. For instance, when asked to fuse ``lentil soup’’ and ``jam’’, \gpt{} repeatedly generated variations of lentil soup served alongside a separately prepared jam, plated in the same way (e.g., placing a dollop of jam in the center of the soup). This fixation persisted when asked to combine specific given recipes, and even when asked to merge broader dish types (e.g., a general soup with a general dip). 

\begin{figure}[t!]
\includegraphics[width=\linewidth]{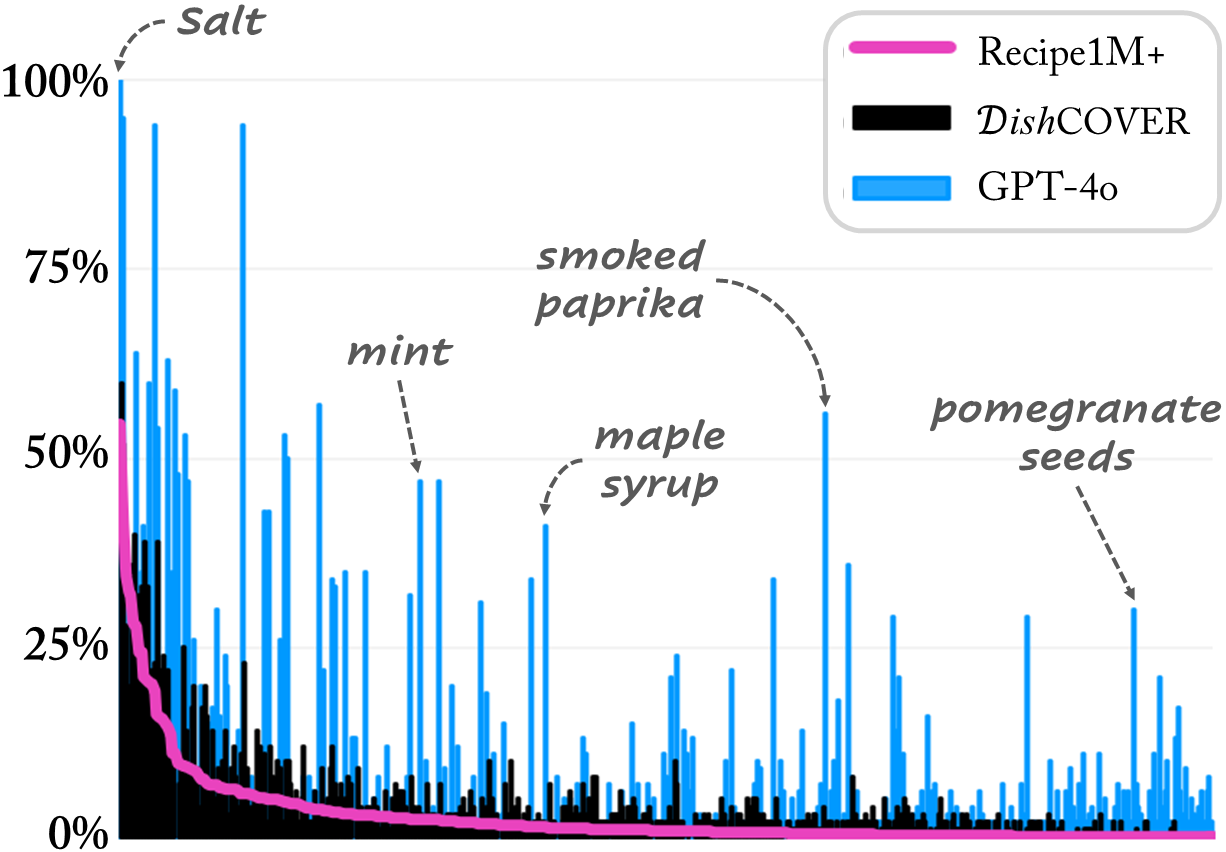}
\caption{\label{fig:ingredient_counts}
Comparing ingredient frequencies in the original recipe repository (in pink) and the models' generated recipes (Exp. 2). The black histogram shows \alg{} closely follows the repository's distribution.   \gpt{} (blue) shows spikes, highlighting bias to certain ingredients. 
}
\end{figure}

Similarly, when asked to suggest a general creative recipe, \gpt{} repeatedly used the same (uncommon) ingredients: 56\% of its recipes included smoked paprika (compared to just 0.375\% in the repository), 47\% used mint, 41\% maple syrup, and 29\% pomegranate molasses. Figure~\ref{fig:ingredient_counts} shows ingredient frequencies in  recipes of \gpt{} and \alg{} against the repository baseline. While \alg{}’s distribution aligns closely with the original data, \gpt{} has a strong bias toward certain low-frequency ingredients.

\paragraph{Quantifying Diversity via Tree Distances.}
We computed the average normalized tree edit distance \cite{rico2003comparison} between generated recipes after converting them into hierarchical tree structures. In both experiments, \gpt{}’s outputs had significantly lower tree distances than \alg{}’s, indicating higher similarity. In the first experiment, we calculated the average edit distance for each dish pair separately, finding that \alg{} had an average tree distance of 132.14, while \gpt{}’s was 89.35 (p-value=3.6e-05, paired t-test). In the second experiment, across all outputs, \alg{} again exhibited greater diversity, with an average tree distance of 140.25 compared to \gpt{}’s 129.55 (p-value<1e-50, two-sample t-test).

\paragraph{Quantifying Lexical Diversity via Self-BLEU.} 
We computed self-BLEU scores across the generated recipes for both experiments. \gpt{} exhibited significantly higher redundancy, reflected in higher self-BLEU scores: 
In Experiment 1, \gpt{}’s outputs scored 0.726 (BLEU-2), 0.655 (BLEU-3), and 0.599 (BLEU-4), while \alg{}’s scores were substantially lower: 0.423, 0.281, and 0.192, respectively. Similarly, in Experiment 2 \gpt{} scored 0.902, 0.828, and 0.753, compared to \alg{}’s 0.778, 0.619, and 0.475. The differences are more pronounced in Experiment 1 due to the smaller reference set (5 recipes per pair, vs.~100 recipes).

\begin{figure}[t!]
\includegraphics[width=.99\linewidth]{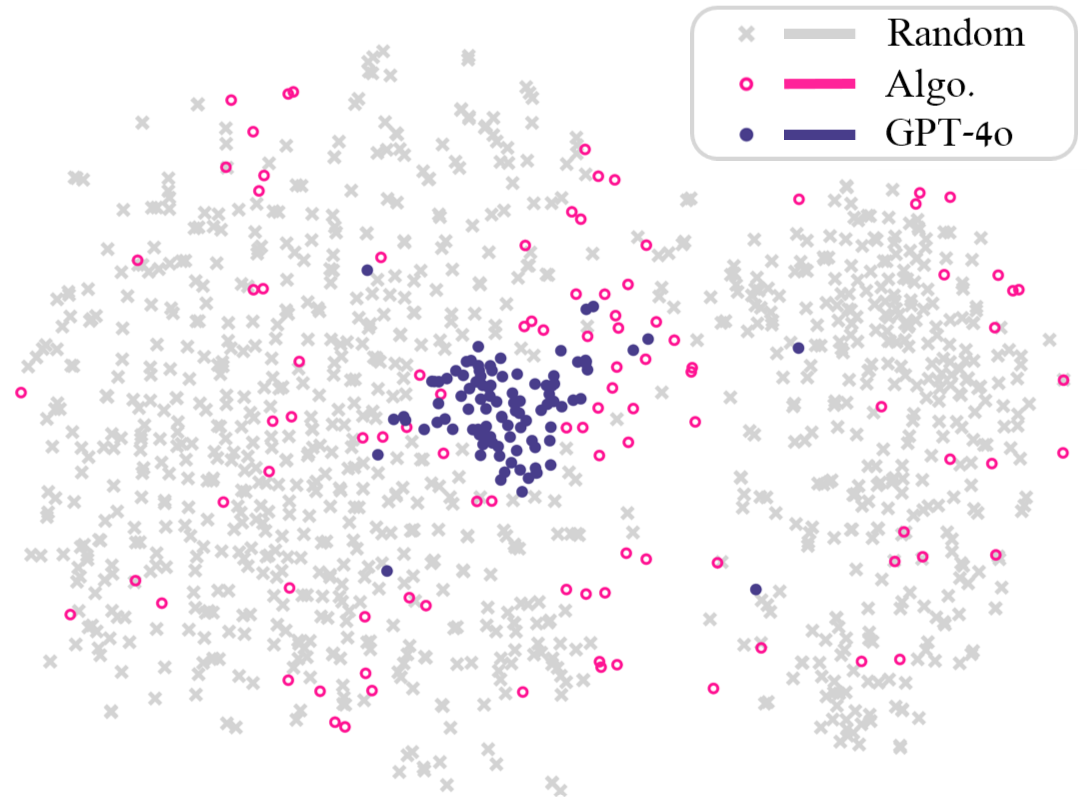}
\caption{\label{fig:tsne_exp2}
t-SNE visualization for recipe embeddings, experiment 2.  The x markers represent 1K random recipes from the general repository. Again,  \alg{}'s outputs are more diverse.
}
\end{figure}

\begin{figure}[t!]
\includegraphics[width=.99\linewidth]{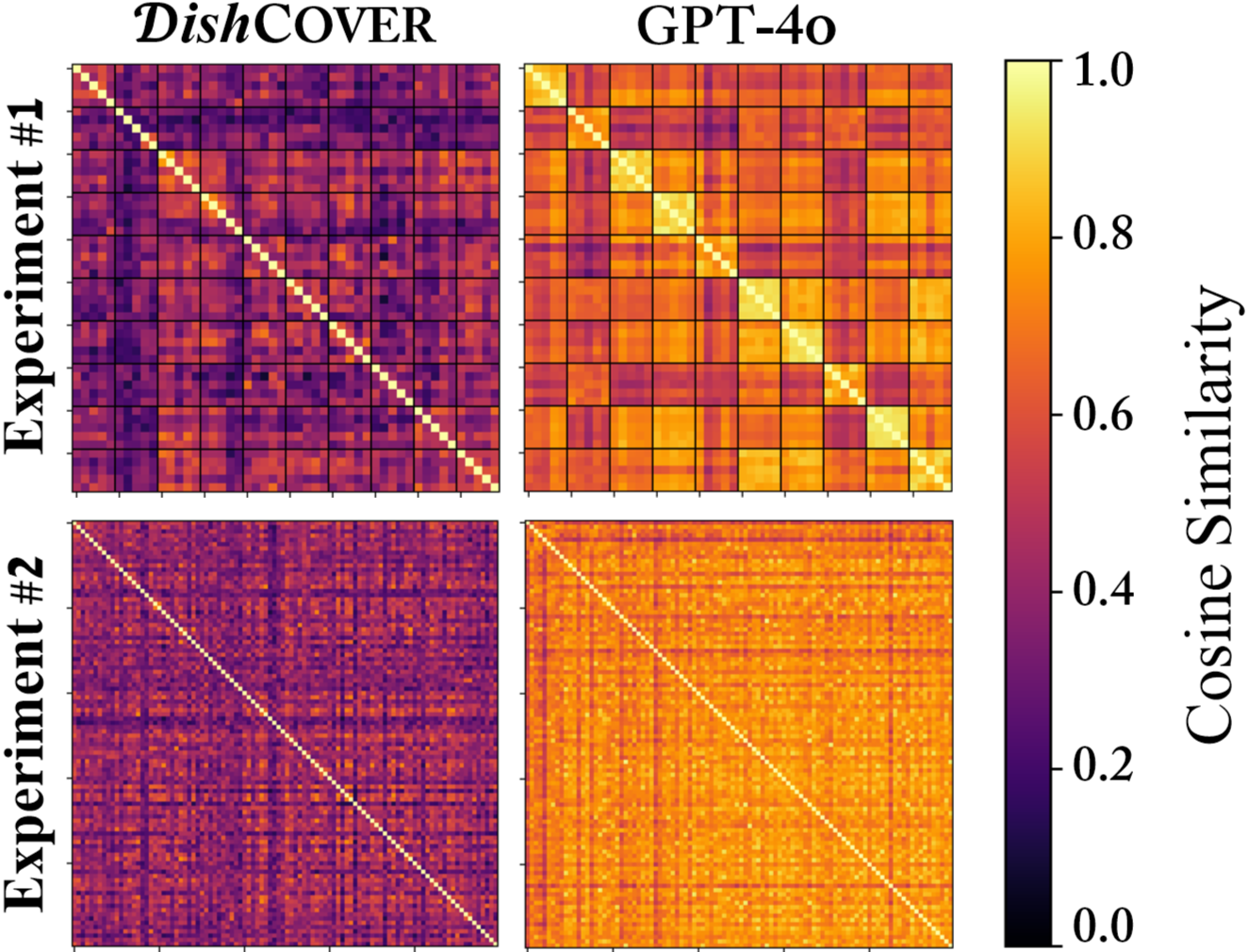}
\caption{\label{fig:heatmaps}
Four heatmaps illustrating pairwise cosine similarities of recipe embeddings for each model across both experiments. \gpt{}’s heatmaps (right) display more areas of high similarity, indicating \alg{}'s higher diversity. 
}
\end{figure}

% \paragraph{Quantifying Diversity via Embeddings.}
\paragraph{Quantifying Intra-Set Diversity.}
We also analyzed the embeddings of the generated recipes using a fine-tuned Sentence-BERT model specialized for cooking recipes (see Appendix~\ref{algo:sbert_finetuned}). Figure~\ref{fig:tsne_exp1} presents a t-SNE visualization for the first experiment, where each shape represents a dish pair, and color indicates the model (\alg{} in pink, \gpt{} in blue). \gpt{}’s recipes formed tight clusters for each pair, whereas \alg{}’s were more dispersed.
A similar pattern emerges in experiment 2 (Figure~\ref{fig:tsne_exp2}). By construction, \alg{}’s embeddings are widely distributed. The interesting thing to notice, however, is that \gpt{}’s embeddings tend to cluster closely.
Figure~\ref{fig:heatmaps} reinforces these findings with heatmaps of cosine similarities among all recipe embeddings. In the first experiment, \alg{}’s outputs had an average similarity of 0.387 (std=0.120), while \gpt{}’s were higher at 0.659 (std=0.121). Results were similar in Experiment 2 (\alg{}: 0.402, std=0.110; \gpt{}: 0.731, std=0.078).

\paragraph{Quantifying Diversity with Respect to an External Corpus.}
We also assessed the similarity of generated recipes  to known recipes by computing cosine similarity between a recipe's embedding and that of its nearest neighbor in the Recipe1M+ corpus.
\alg{}'s recipes consistently exhibited greater distance from existing recipes, with an average similarity of 0.774 (std=0.061), significantly lower than \gpt{}'s average of 0.851 (std=0.027). Additionally, 91\% of \alg{}'s recipes had  similarity below 0.85, compared to only 47\% of \gpt{}'s. Notably, 38\% of \alg{}'s outputs had similarity score below 0.75, while none of \gpt{}'s did.

\begin{figure}[t!]
\includegraphics[width=\linewidth]{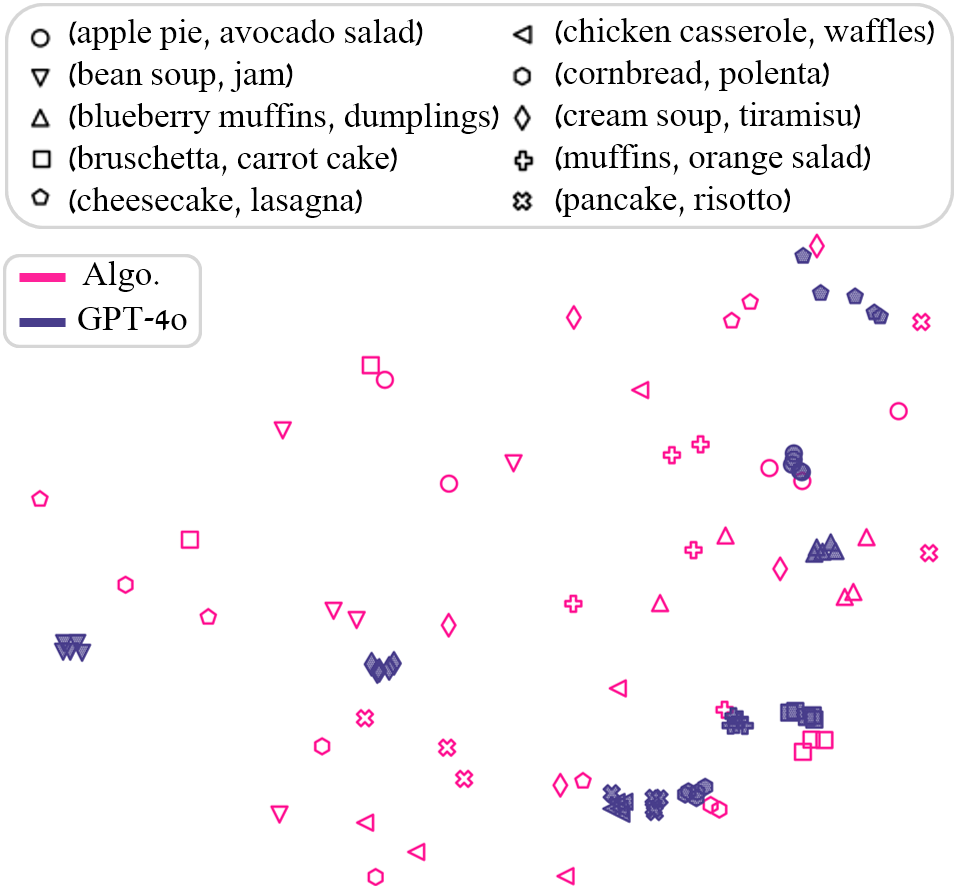}
\caption{\label{fig:tsne_exp1}
t-SNE visualization for recipe embeddings, experiment 1. Shapes denote dish pairs, colors and fill style denote models. The filled clusters (\gpt{}) of the same shape appear tight and localized, while the unfilled markers (\alg{}) are scattered, showing higher diversity across \alg{}'s outputs. 
}
\end{figure}

% \begin{table}[t!]
% \centering
% \renewcommand{\arraystretch}{1.3}
% \SMALL
% \resizebox{\columnwidth}{!}{%
% \begin{tabular}{@{}lcccc@{}}
% \toprule
% &  & \textbf{BLEU-2} & \textbf{BLEU-3} & \textbf{BLEU-4}
% \\ \midrule
% \textbf{\textsc{EXP\#1}} & \cgpt{}  & .726 & .655  & .599\\ 
% & \alg{}  & .423 & .281  & .192\\
% \midrule
% \textbf{\textsc{EXP\#2}} & \cgpt{}  & .902 & .828  & .753\\
% & \alg{}  & .778 & .619  & .475\\
%  \bottomrule
% \end{tabular}%
% }
% \caption{self-BLEU values for both experiments. 
% }
% \label{tab:self_bleu_exp1}
% \end{table}

\subsection{Human annotation}

While diversity is crucial, genuine creativity in recipe generation also depends on value and novelty.
To assess these aspects, we built on the cooking expert annotations from the human evaluation.

In terms of \textbf{\textit{value}}, both models produced outputs that were largely deemed valuable. In the first experiment, 80\% of \alg{}’s recipes were classified as valuable, with an average value score of 4.320, compared to 82\% for \gpt{} (4.326). In the second experiment, 85\% of \alg{}’s outputs were classified as valuable (4.36), compared to 98\% for \gpt{} (4.51). This makes sense, as we noted that the second experiment (open-ended prompts) is more similar to the data \gpt{} has encountered during training, and also that it tends to output relatively similar recipes. 

For \textbf{\textit{novelty}}, we considered only the recipes that were classified as valuable; this reflects a use-case of a cook skimming a list of recommended recipes, quickly judging their sensibility, delving only into those that have potential. In both experiments, \alg{} had significantly outperformed \gpt{} in this regard. In the first experiment, \alg{}’s average novelty score was 3.53, compared to \gpt{}’s 3.146 (p-value=0.0009). In the second experiment, \alg{}’s average novelty score was 3.612, compared to \gpt{}’s 3.141 (p-value=9.2E-08).

Examining the second experiment’s valuable outputs further, \alg{} dominated the top quartile of novelty scores, making up 75.55\% of the highest-rated recipes, while \gpt{} prevailed in the lower two quartiles, accounting for 71.74\% of the less novel outputs. 
Moreover, among the 37 valuable recipes with a novelty score of 4 or higher, \alg{} contributed 32, leaving only five from \gpt{}. Figure~\ref{fig:main} showcases five \alg{} recipes from this high-novelty set. These results strongly suggest that while both models produce mostly valuable recipes (\gpt{} more so, in the open-ended case), \alg{} has a clear advantage in generating truly creative ones.

% \newpage
\section{Related Work}
\label{sec:related_work}
% LLMs have demonstrated remarkable fluency and coherence in text generation. However, they often fall short when it comes to producing truly novel ideas \cite{franceschelli2024creativity, chakrabarty2024art, tian2024large, zhao2024assessing}. Their outputs tend to be repetitive \cite{jentzsch2023chatgpt}, lacking in diversity \cite{kirk2023understanding,zhao2025large, ma2025large}, and heavily shaped by the statistical patterns learned during training \ \cite{lee2023language, chang2024language}. As a result, LLMs frequently generate content that mirrors existing knowledge rather than introducing genuinely creative or original concepts. This reliance on seen data also raises concerns about memorization and potential plagiarism \cite{chang2024language, lee2023language}. These challenges are further compounded by the difficulty of measuring creativity, which has been widely studied in humans, computational systems, and more recently, in LLMs \cite{said2017approaches, lamb2018evaluating, chakrabarty2024art, zhao2024assessing}.

Our approach builds on recent parsing-based methods that guide LLMs to map natural language into structured forms, boosting LLM performance in tasks such as constituency parsing \cite{tian2024largel} and information extraction \cite{zhao2023large, li2024simple}. 
% Similarly, we use LLMs to transform domain-specific input into structured forms, enhancing adaptability across tasks.
%
Similar to research integrating LLMs with Knowledge Graphs (KGs) to improve inference and reasoning \cite{feng2023knowledge, jiang2023structgpt, sun2023think, wang2024knowledge}, we incorporate domain-specific knowledge into structured representations to provide LLMs with clearer, context-rich signals.

Our work also aligns with models that parse text into structured knowledge, manipulate these representations, and (optionally) convert them into natural language to enhance LLM capabilities \cite{yang2023coupling,zelikman2023parsel,besta2024graph,zhang2025sr}. 
% While previous works primarily focused on improving inference and reasoning, 
We surprisingly show that similar techniques can augment creativity and diversity.
In parallel, the structured representations we use share a conceptual similarity with latent variable models, which learn to encode inputs into structured latent forms and apply transformations to generate viable alternatives \cite{kusner2017grammar, dai2018syntax, zhang2019syntax}. 
Unlike these approaches, our method does not learn a latent space and requires no training.
%, adapting these structured processing techniques to enhance generative flexibility.

Our focus on increasing LLM output diversity complements efforts that promote variation in generation via human feedback \cite{chung2023increasing}, in-context learning \cite{zhang2024improving}, or KG-based interventions 
\cite{liu2021kg, hwang2023knowledge, liu2022dimongen}. Additionally, our work aligns with broader research aimed at fostering creativity in LLMs.
%
% To address this gap, recent research has explored various methods to enhance creativity of LLMs. 
One line of work leverages hallucinations for new ideas \cite{jiang2024survey, yuan2025hallucinations}.
%, using the model's ability to generate imaginative but sometimes incorrect outputs . 
Another one draws on insights from human creativity research, incorporating techniques such as constraints \cite{ lu2024benchmarking}, associative thinking \cite{mehrotra2024enhancing}, role-playing \cite{ chen2024hollmwood}, 
and brainstorming \cite{summers2023brainstorm, chang2025framework, rana2025generative}. 
Similarly, we focus on recombination.
% -- the process of generating new ideas by meaningfully combining existing ones. 

Common computational approaches for recombining ideas include conceptual blending 
frameworks, where merging is guided by heuristics or rules \cite{fauconnier2003conceptual, pereira2006experiments, veale2019computational};
% camara2007creativity}; 
genetic algorithms, which use random crossover points \cite{corne2001creative, cho2002towards, dennis2011teaching}; and latent space blending, which interpolates between learned representations but lacks explicit structure \cite{sarkar2021generating, yee2022latent, zhou2025freeblend}. 
% Our approach differs by recombining structured representations using an edit distance algorithm, which provides explicit and controllable structural blending.
Our approach differs by recombining structured representations through an edit distance algorithm, thereby providing explicit and controllable blending.

Beyond improving LLMs, researchers have also explored their use as creative aids for writers \cite{yuan2022wordcraft,mirowski2023co,chakrabarty2023creativity,
wan2024felt}, 
% , 
visual artists \cite{ko2023large}, and even humorists \cite{wu2025one}. However, while these tools boost users' sense of creativity, their limited diversity may homogenize the ideas produced across different individuals \cite{anderson2024homogenization}.

% LLMs were explored in the culinary domain \cite{ma2024large, zhou2024foodsky, ataguba2025exploring, 
% hwang2023large}, also for recipe generation \cite{h2020recipegpt, anto2020creative, bien2020cooking}. Pre-LLM work focused on ingredient combinations, often neglecting instruction generation \cite{morris2012soup, amorim2017creative, cromwell2015computational, varshney2019big}.

Recent work has explored the use of LLMs in the culinary domain \cite{hwang2023large,ma2024large, zhou2024foodsky, ataguba2025exploring, thomascan2025}, including several early efforts in recipe generation \cite{h2020recipegpt, anto2020creative, bien2020cooking}. Prior to the advent of LLMs, computational recipe generation focused primarily on proposing novel ingredient combinations, often neglecting complete cooking instruction generation \cite{morris2012soup,cromwell2015computational,amorim2017creative,varshney2019big}.

% The project most aligned with our goals is Chef Watson \cite{varshney2019big}, an early system that generated new recipes by modeling aspects of human creativity. However, it was highly domain-specific, focusing on ingredient combinations without broader abstraction and predating modern LLMs. In contrast, we use an LLM to translate recipes into a mathematical representation, enabling recombination at a fundamental level. We believe this abstraction makes our framework more adaptable, extending beyond cooking to support creativity across diverse domains.

\section{Discussion \& Future Work}
\label{sec:discussion}
\xhdr{Generalizability to Other Domains}
% \subsection*{Generalizability \& Domain-Specific Considerations}
Extending our approach beyond the culinary domain presents important considerations. Our pipeline fundamentally relies on (1) meaningful \textbf{structured representations}, and (2) the ability to define or approximate \textbf{reliable criteria for assessing value}.

As discussed in Section~\ref{sec:algorithm}, procedural texts naturally exhibit tree-like structures: leaves often correspond to objects, and internal nodes represent operations performed on them. In storytelling,  graphs are commonly used to model narrative arcs and plot dynamics \cite{elson2012modeling, valls2017towards}. In domains such as drug discovery, molecular structures are inherently graph-based, with atoms as nodes and bonds as edges \cite{ivanciuc2000graph}. Music also allows structured, hierarchical representations  \cite{good2012musicxml, cuthbert2010music21}. Other domains, such as poetry or product ideation, lack obvious or standardized structured representations, making the applicability of our technique less immediate.

Value assessment poses another challenge. In recipes, we leverage the FlavorDB dataset as a proxy metric for taste. Our method can be adapted to other domains that employ similar domain-specific metrics. For example, in music, one can predict how humans perceive a combination of sounds (based on psychoacoustic consonance heuristics). Domains related to physical construction (furniture, architecture) could apply automated feasibility checks through CAD software. For programming and game design, functionality could similarly be verified automatically through simulation. In domains where value assessments are inherently subjective and difficult to automate, domain expert evaluation could be integrated into the pipeline, enabling hybrid approaches.
%that balance structured generation with  expert-driven assessment.}

%\madd{
\xhdr{Structured Recombination 
% for High-Level Variability 
in LLMs: Implications for AI Creativity and Sampling}
A core insight of our work is that structured recombination provides a controllable mechanism to introduce meaningful variability at a higher level of abstraction, rather than the lexical, token level.

%\madd{\xhdr{Implications for AI Creativity} 
To date, attempts to enhance creativity in LLMs often focused on increasing token-randomness (e.g., via temperature), which is not sufficient for creativity \cite{peeperkorn2024temperature}. Structured recombination enables a model to explore the creative space of ideas more deliberately and effectively. Together with modules for assessing value and novelty this can lead to more creative outputs. %novel, yet valuable outputs.

%\madd{\xhdr{Implications for Sampling}
Another promising application of structured recombination is in {\bf structured sampling} and search-based generation tasks. Standard sampling methods in LLMs \cite{fan2018hierarchical, holtzman2019curious} often rely on stochastic techniques that operate at the token level (e.g., nucleus or top-k sampling) to introduce variability. In contrast, we diversify by sampling over abstract representations rather than raw tokens. We introduce a controlled yet flexible way of navigating idea spaces, which can enhance a range of applications that require deliberate exploration of diverse concepts. % and solutions. , previously unexplored

\medskip
\section{Conclusions}
In this work, we addressed the persistent challenge of generating creative, diverse outputs. We proposed a novel approach that leverages structured representations to enhance creativity. Rather than relying on superficial token-level variation, we perform cognitively-inspired manipulations -- specifically, recombining structured representations of existing concepts. We demonstrated our paradigm in the culinary domain through \alg{}, a model designed to generate creative and diverse recipes, and empirically showed significant improvements over \cgpt{} in terms of creativity.
Ultimately, our approach represents a step toward pushing LLMs beyond surface-level variation, opening the path to richer and more controllable creative generation. We hope this work inspires further research into structured creativity in AI, and we invite the community to build upon this paradigm across diverse domains.

\section*{Acknowledgements}
% ACKWONLEDGEMENTS
We thank the reviewers and action editor for their insightful comments. We further thank Dana Aviran and Eitan Stern for developing the Sentence-BERT fine-tuned model on recipe data, and to the members of \textbf{\href{https://www.hyadatalab.com/}{Hyadata Lab}} and \textbf{\href{https://gabrielstanovsky.github.io/group/}{SLAB}} at the Hebrew University of Jerusalem for their thoughtful remarks. This work was supported by the European Research Council (ERC) under the European Union's Horizon 2020 research and innovation programme (grant no. 852686, SIAM), by the Israeli Ministry of Science and Technology (grant no. 7256), and by the Koret Foundation grant for Smart Cities and Digital Living.

%\section{Conclusions}
%\label{sec:conclusion and Future Work}
%\input{sections/08_conclusion}

% \section*{Acknowledgements}
% \input{sections/acknowledgements}

\bibliography{tacl2021}

\begin{thebibliography}{103}
\expandafter\ifx\csname natexlab\endcsname\relax\def\natexlab#1{#1}\fi

\bibitem[{Ahuja and Morris~Lampert(2001)}]{ahuja2001entrepreneurship}
Gautam Ahuja and Curba Morris~Lampert. 2001.
\newblock Entrepreneurship in the large corporation: A longitudinal study of how established firms create breakthrough inventions.
\newblock \emph{Strategic management journal}, 22(6-7):521--543.

\bibitem[{Amorim et~al.(2017)Amorim, G{\'o}es, Da~Silva, and Fran{\c{c}}a}]{amorim2017creative}
Alvaro Amorim, Lu{\'\i}s~FW G{\'o}es, Alysson~Ribeiro Da~Silva, and Celso Fran{\c{c}}a. 2017.
\newblock Creative flavor pairing: Using {R}{D}{C} metric to generate and assess ingredients combination.
\newblock In \emph{ICCC}, pages 33--40.

\bibitem[{Anderson et~al.(2024)Anderson, Shah, and Kreminski}]{anderson2024homogenization}
Barrett~R. Anderson, Jash~Hemant Shah, and Max Kreminski. 2024.
\newblock Homogenization effects of large language models on human creative ideation.
\newblock In \emph{Proceedings of the 16th conference on creativity \& cognition}, pages 413--425.

\bibitem[{Ant{\^o}nio et~al.(2020)Ant{\^o}nio, Bezerra, G{\'o}es, and Ferreira}]{anto2020creative}
Willian Ant{\^o}nio, Jo{\~a}o~Ribeiro Bezerra, Lu{\'\i}s Fabr{\'\i}cio~Wanderley G{\'o}es, and Fl{\'a}via Magalh{\~a}es~Freitas Ferreira. 2020.
\newblock Creative culinary recipe generation based on statistical language models.
\newblock \emph{IEEE Access}, 8:146263--146283.

\bibitem[{Arora et~al.(1998)Arora, Lund, Motwani, Sudan, and Szegedy}]{arora1998proof}
Sanjeev Arora, Carsten Lund, Rajeev Motwani, Madhu Sudan, and Mario Szegedy. 1998.
\newblock Proof verification and the hardness of approximation problems.
\newblock \emph{Journal of the ACM (JACM)}, 45(3):501--555.

\bibitem[{Ataguba and Orji(2025)}]{ataguba2025exploring}
Grace Ataguba and Rita Orji. 2025.
\newblock Exploring large language models for personalized recipe generation and weight-loss management.
\newblock \emph{ACM Transactions on Computing for Healthcare}.

\bibitem[{Bertsimas and Tsitsiklis(1993)}]{bertsimas1993simulated}
Dimitris Bertsimas and John Tsitsiklis. 1993.
\newblock Simulated annealing.
\newblock \emph{Statistical science}, 8(1):10--15.

\bibitem[{Besta et~al.(2024)Besta, Blach, Kubicek, Gerstenberger, Podstawski, Gianinazzi, Gajda, Lehmann, Niewiadomski, Nyczyk, and Hoefler}]{besta2024graph}
Maciej Besta, Nils Blach, Ales Kubicek, Robert Gerstenberger, Michal Podstawski, Lukas Gianinazzi, Joanna Gajda, Tomasz Lehmann, Hubert Niewiadomski, Piotr Nyczyk, and Torsten Hoefler. 2024.
\newblock Graph of thoughts: Solving elaborate problems with large language models.
\newblock In \emph{Proceedings of the AAAI Conference on Artificial Intelligence}, volume~38, pages 17682--17690.

\bibitem[{Bie{\'n} et~al.(2020)Bie{\'n}, Gilski, Maciejewska, Taisner, Wisniewski, and Lawrynowicz}]{bien2020cooking}
Micha{\l} Bie{\'n}, Micha{\l} Gilski, Martyna Maciejewska, Wojciech Taisner, Dawid Wisniewski, and Agnieszka Lawrynowicz. 2020.
\newblock Recipe{N}{L}{G}: A cooking recipes dataset for semi-structured text generation.
\newblock \emph{Proceedings of the 13th international conference on natural language generation}, pages 22--28.

\bibitem[{Bille(2005)}]{bille2005survey}
Philip Bille. 2005.
\newblock A survey on tree edit distance and related problems.
\newblock \emph{Theoretical computer science}, 337(1-3):217--239.

\bibitem[{Boden(2004)}]{boden2004creative}
Margaret~A. Boden. 2004.
\newblock \emph{The Creative Mind: Myths and Mechanisms}.
\newblock Routledge.

\bibitem[{Boden(2009)}]{boden2009computer}
Margaret~A. Boden. 2009.
\newblock Computer models of creativity.
\newblock \emph{AI Magazine}, 30(3):23--23.

\bibitem[{Chakrabarty et~al.(2024{\natexlab{a}})Chakrabarty, Laban, Agarwal, Muresan, and Wu}]{chakrabarty2024art}
Tuhin Chakrabarty, Philippe Laban, Divyansh Agarwal, Smaranda Muresan, and Chien-Sheng Wu. 2024{\natexlab{a}}.
\newblock Art or artifice? large language models and the false promise of creativity.
\newblock In \emph{Proceedings of the 2024 CHI Conference on Human Factors in Computing Systems}, pages 1--34.

\bibitem[{Chakrabarty et~al.(2024{\natexlab{b}})Chakrabarty, Padmakumar, Brahman, and Muresan}]{chakrabarty2023creativity}
Tuhin Chakrabarty, Vishakh Padmakumar, Faeze Brahman, and Smaranda Muresan. 2024{\natexlab{b}}.
\newblock Creativity support in the age of large language models: An empirical study involving professional writers.
\newblock In \emph{Proceedings of the 16th Conference on Creativity \& Cognition}, pages 132--155.

\bibitem[{Chang and Li(2025)}]{chang2025framework}
Hung-Fu Chang and Tong Li. 2025.
\newblock A framework for collaborating a large language model tool in brainstorming for triggering creative thoughts.
\newblock \emph{Thinking Skills and Creativity}, page 101755.

\bibitem[{Chen et~al.(2024)Chen, Zhu, Yang, Shi, Xi, Zhang, Wang, Pu, Feng, Yang, and Zhang}]{chen2024hollmwood}
Jing Chen, Xinyu Zhu, Cheng Yang, Chufan Shi, Yadong Xi, Yuxiang Zhang, Junjie Wang, Jiashu Pu, Tian Feng, Yujiu Yang, and Rongsheng Zhang. 2024.
\newblock Ho{L}{L}{M}wood: Unleashing the creativity of large language models in screenwriting via role playing.
\newblock In \emph{Findings of the Association for Computational Linguistics: EMNLP 2024}, pages 8075--8121.

\bibitem[{Cho(2002)}]{cho2002towards}
Sung-Bae Cho. 2002.
\newblock Towards creative evolutionary systems with interactive genetic algorithm.
\newblock \emph{Applied Intelligence}, 16:129--138.

\bibitem[{Chung et~al.(2023)Chung, Kamar, and Amershi}]{chung2023increasing}
John Chung, Ece Kamar, and Saleema Amershi. 2023.
\newblock Increasing diversity while maintaining accuracy: Text data generation with large language models and human interventions.
\newblock In \emph{Proceedings of the 61st Annual Meeting of the Association for Computational Linguistics (Volume 1: Long Papers)}, pages 575--593.

\bibitem[{Corne and Bentley(2001)}]{corne2001creative}
David~W. Corne and Peter~J. Bentley. 2001.
\newblock \emph{Creative Evolutionary Systems}.
\newblock Elsevier.

\bibitem[{Cromwell et~al.(2015)Cromwell, Galeota-Sprung, and Ramanujan}]{cromwell2015computational}
Erol Cromwell, Jonah Galeota-Sprung, and Raghuram Ramanujan. 2015.
\newblock Computational creativity in the culinary arts.
\newblock In \emph{FLAIRS}, pages 38--42.

\bibitem[{Cuthbert and Ariza(2010)}]{cuthbert2010music21}
Michael~Scott Cuthbert and Christopher Ariza. 2010.
\newblock music21: A toolkit for computer-aided musicology and symbolic music data.
\newblock In \emph{11th International Society for Music Information Retrieval Conference (ISMIR 2010)}, pages 637--642.

\bibitem[{Dai et~al.(2018)Dai, Tian, Dai, Skiena, and Song}]{dai2018syntax}
Hanjun Dai, Yingtao Tian, Bo~Dai, Steven Skiena, and Le~Song. 2018.
\newblock Syntax-directed variational autoencoder for structured data.
\newblock In \emph{International Conference on Learning Representations}.

\bibitem[{Dennis and Stella(2011)}]{dennis2011teaching}
John~L. Dennis and Aldo Stella. 2011.
\newblock Teaching creativity: The case for/against genetic algorithms as a model of human creativity.
\newblock \emph{The Open Educational Journal}, 4(1):36--40.

\bibitem[{Doboli et~al.(2020)Doboli, Kenworthy, Paulus, Minai, and Doboli}]{doboli2020cognitive}
Simona Doboli, Jared Kenworthy, Paul Paulus, Ali Minai, and Alex Doboli. 2020.
\newblock A cognitive inspired method for assessing novelty of short-text ideas.
\newblock In \emph{2020 International Joint Conference on Neural Networks (IJCNN)}, pages 1--8. IEEE.

\bibitem[{Elson(2012)}]{elson2012modeling}
David~K. Elson. 2012.
\newblock \emph{Modeling Narrative Discourse}.
\newblock Columbia University.

\bibitem[{Fan et~al.(2018)Fan, Lewis, and Dauphin}]{fan2018hierarchical}
Angela Fan, Mike Lewis, and Yann Dauphin. 2018.
\newblock Hierarchical neural story generation.
\newblock In \emph{Proceedings of the 56th Annual Meeting of the Association for Computational Linguistics (Volume 1: Long Papers)}, pages 889--898.

\bibitem[{Fauconnier and Turner(2003)}]{fauconnier2003conceptual}
Gilles Fauconnier and Mark Turner. 2003.
\newblock Conceptual blending, form and meaning.
\newblock \emph{Recherches en communication}, 19:57--86.

\bibitem[{Feng et~al.(2023)Feng, Zhang, and Fei}]{feng2023knowledge}
Chao Feng, Xinyu Zhang, and Zichu Fei. 2023.
\newblock Knowledge solver: Teaching {L}{L}{M}s to search for domain knowledge from knowledge graphs.
\newblock \emph{arXiv preprint arXiv:2309.03118v1}.

\bibitem[{Finke et~al.(1996)Finke, Ward, and Smith}]{finke1996creative}
Ronald~A. Finke, Thomas~B. Ward, and Steven~M. Smith. 1996.
\newblock \emph{Creative Cognition: Theory, Research, and Applications}.
\newblock MIT press.

\bibitem[{Franceschelli and Musolesi(2024)}]{franceschelli2024creativity}
Giorgio Franceschelli and Mirco Musolesi. 2024.
\newblock On the creativity of large language models.
\newblock \emph{AI \& SOCIETY}, pages 1--11.

\bibitem[{Fukagawa et~al.(2022)Fukagawa, McKillop, Pehrsson, Moshfegh, Harnly, and Finley}]{fukagawa2022usda}
Naomi~K. Fukagawa, Kyle McKillop, Pamela~R. Pehrsson, Alanna Moshfegh, James Harnly, and John Finley. 2022.
\newblock {U}{S}{D}{A}’s {F}ood{D}ata {C}entral: {W}hat is it and why is it needed today?
\newblock \emph{The American journal of clinical nutrition}, 115(3):619--624.

\bibitem[{Garg et~al.(2018)Garg, Sethupathy, Tuwani, Nk, Dokania, Iyer, Gupta, Agrawal, Singh, Shukla, Kathuria, Badhwar, Kanji, Jain, Kaur, Nagpal, and Bagler}]{garg2018flavordb}
Neelansh Garg, Apuroop Sethupathy, Rudraksh Tuwani, Rakhi Nk, Shubham Dokania, Arvind Iyer, Ayushi Gupta, Shubhra Agrawal, Navjot Singh, Shubham Shukla, Kriti Kathuria, Rahul Badhwar, Rakesh Kanji, Anupam Jain, Avneet Kaur, Rashmi Nagpal, and Ganesh Bagler. 2018.
\newblock {F}lavor{D}{B}: A database of flavor molecules.
\newblock \emph{Nucleic acids research}, 46(D1):D1210--D1216.

\bibitem[{Good(2001)}]{good2012musicxml}
Michael Good. 2001.
\newblock {M}usic{X}{M}{L} for notation and analysis.
\newblock \emph{The virtual score: representation, retrieval, restoration}, 12(113-124):160.

\bibitem[{Guilford(1967)}]{guilford1967nature}
Joy~Paul Guilford. 1967.
\newblock \emph{The Nature of Human Intelligence.}
\newblock McGraw-Hill.

\bibitem[{H.~Lee et~al.(2020)H.~Lee, Shu, Achananuparp, Prasetyo, Liu, Lim, and Varshney}]{h2020recipegpt}
Helena H.~Lee, Ke~Shu, Palakorn Achananuparp, Philips~Kokoh Prasetyo, Yue Liu, Ee-Peng Lim, and Lav~R. Varshney. 2020.
\newblock Recipe{G}{P}{T}: Generative pre-training based cooking recipe generation and evaluation system.
\newblock In \emph{Companion Proceedings of the Web Conference 2020}, pages 181--184.

\bibitem[{Heinen and Johnson(2018)}]{heinen2018semantic}
David~J.P. Heinen and Dan~R. Johnson. 2018.
\newblock Semantic distance: An automated measure of creativity that is novel and appropriate.
\newblock \emph{Psychology of Aesthetics, Creativity, and the Arts}, 12(2):144.

\bibitem[{Holtzman et~al.(2019)Holtzman, Buys, Du, Forbes, and Choi}]{holtzman2019curious}
Ari Holtzman, Jan Buys, Li~Du, Maxwell Forbes, and Yejin Choi. 2019.
\newblock The curious case of neural text degeneration.
\newblock In \emph{International Conference on Learning Representations}.

\bibitem[{Hwang et~al.(2023{\natexlab{a}})Hwang, Li, Hou, and Roth}]{hwang2023large}
Alyssa Hwang, Bryan Li, Zhaoyi Hou, and Dan Roth. 2023{\natexlab{a}}.
\newblock Large language models as sous chefs: Revising recipes with {G}{P}{T}-3.
\newblock \emph{arXiv preprint arXiv:2306.13986v1}.

\bibitem[{Hwang et~al.(2023{\natexlab{b}})Hwang, Thost, Shwartz, and Ma}]{hwang2023knowledge}
EunJeong Hwang, Veronika Thost, Vered Shwartz, and Tengfei Ma. 2023{\natexlab{b}}.
\newblock Knowledge graph compression enhances diverse commonsense generation.
\newblock In \emph{Proceedings of the 2023 Conference on Empirical Methods in Natural Language Processing}, pages 558--572.

\bibitem[{Ivanciuc and Balaban(2000)}]{ivanciuc2000graph}
Ovidiu Ivanciuc and Alexandru~T. Balaban. 2000.
\newblock The graph description of chemical structures.
\newblock In \emph{Topological indices and related descriptors in QSAR and QSPR}, pages 69--178. CRC Press.

\bibitem[{Jermsurawong and Habash(2015)}]{jermsurawong2015predicting}
Jermsak Jermsurawong and Nizar Habash. 2015.
\newblock Predicting the structure of cooking recipes.
\newblock In \emph{Proceedings of the 2015 Conference on Empirical Methods in Natural Language Processing}, pages 781--786.

\bibitem[{Jiang et~al.(2023)Jiang, Zhou, Dong, Ye, Zhao, and Wen}]{jiang2023structgpt}
Jinhao Jiang, Kun Zhou, Zican Dong, Keming Ye, Wayne~Xin Zhao, and Ji-Rong Wen. 2023.
\newblock Struct{G}{P}{T}: A general framework for large language model to reason over structured data.
\newblock In \emph{Proceedings of the 2023 Conference on Empirical Methods in Natural Language Processing}, pages 9237--9251.

\bibitem[{Jiang et~al.(2024)Jiang, Tian, Hua, Xu, Wang, and Guo}]{jiang2024survey}
Xuhui Jiang, Yuxing Tian, Fengrui Hua, Chengjin Xu, Yuanzhuo Wang, and Jian Guo. 2024.
\newblock A survey on large language model hallucination via a creativity perspective.
\newblock \emph{arXiv preprint arXiv:2402.06647v1}.

\bibitem[{Jordanous(2012)}]{jordanous2012evaluating}
Anna~Katerina Jordanous. 2012.
\newblock \emph{Evaluating Computational Creativity: A Standardised Procedure for Evaluating Creative Systems and Its Application}.
\newblock University of Kent (United Kingdom).

\bibitem[{Kenett(2019)}]{kenett2019can}
Yoed~N. Kenett. 2019.
\newblock What can quantitative measures of semantic distance tell us about creativity?
\newblock \emph{Current Opinion in Behavioral Sciences}, 27:11--16.

\bibitem[{Ko et~al.(2023)Ko, Park, Jeon, Jo, Kim, and Seo}]{ko2023large}
Hyung-Kwon Ko, Gwanmo Park, Hyeon Jeon, Jaemin Jo, Juho Kim, and Jinwook Seo. 2023.
\newblock Large-scale text-to-image generation models for visual artists’ creative works.
\newblock In \emph{Proceedings of the 28th international conference on intelligent user interfaces}, pages 919--933.

\bibitem[{Koestler(1964)}]{koestler1964act}
Arthur Koestler. 1964.
\newblock \emph{The Act of Creation}.
\newblock London Hutchinson.

\bibitem[{Kusner et~al.(2017)Kusner, Paige, and Hern{\'a}ndez-Lobato}]{kusner2017grammar}
Matt~J. Kusner, Brooks Paige, and Jos{\'e}~Miguel Hern{\'a}ndez-Lobato. 2017.
\newblock Grammar variational autoencoder.
\newblock In \emph{International conference on machine learning}, pages 1945--1954. PMLR.

\bibitem[{Lamb et~al.(2018)Lamb, Brown, and Clarke}]{lamb2018evaluating}
Carolyn Lamb, Daniel~G. Brown, and Charles~L.A. Clarke. 2018.
\newblock Evaluating computational creativity: An interdisciplinary tutorial.
\newblock \emph{ACM Computing Surveys (CSUR)}, 51(2):1--34.

\bibitem[{Li et~al.(2024)Li, Ramprasad, and Zhang}]{li2024simple}
Yinghao Li, Rampi Ramprasad, and Chao Zhang. 2024.
\newblock A simple but effective approach to improve structured language model output for information extraction.
\newblock In \emph{Findings of the Association for Computational Linguistics: EMNLP 2024}, pages 5133--5148.

\bibitem[{Liu et~al.(2023)Liu, Huang, Zhu, and Chang}]{liu2022dimongen}
Chenzhengyi Liu, Jie Huang, Kerui Zhu, and Kevin Chen-Chuan Chang. 2023.
\newblock Dimon{G}en: Diversified generative commonsense reasoning for explaining concept relationships.
\newblock In \emph{Proceedings of the 61st Annual Meeting of the Association for Computational Linguistics (Volume 1: Long Papers)}, pages 4719--4731.

\bibitem[{Liu et~al.(2021)Liu, Wan, He, Peng, and Yu}]{liu2021kg}
Ye~Liu, Yao Wan, Lifang He, Hao Peng, and Philip~S. Yu. 2021.
\newblock {K}{G}-{B}{A}{R}{T}: Knowledge graph-augmented {B}{A}{R}{T} for generative commonsense reasoning.
\newblock In \emph{Proceedings of the AAAI conference on artificial intelligence}, volume~35, pages 6418--6425.

\bibitem[{Lu et~al.(2025)Lu, Wang, Li, Jiang, Khudanpur, Jiang, and Khashabi}]{lu2024benchmarking}
Yining Lu, Dixuan Wang, Tianjian Li, Dongwei Jiang, Sanjeev Khudanpur, Meng Jiang, and Daniel Khashabi. 2025.
\newblock Benchmarking language model creativity: A case study on code generation.
\newblock In \emph{Proceedings of the 2025 Conference of the Nations of the Americas Chapter of the Association for Computational Linguistics: Human Language Technologies (Volume 1: Long Papers)}, pages 2776--2794.

\bibitem[{Ma et~al.(2024)Ma, Tsai, He, Jia, Zhen, Yu, Wang, Ahuja, and Wei}]{ma2024large}
Peihua Ma, Shawn Tsai, Yiyang He, Xiaoxue Jia, Dongyang Zhen, Ning Yu, Qin Wang, Jaspreet~K.C. Ahuja, and Cheng-I Wei. 2024.
\newblock Large language models in food science: Innovations, applications, and future.
\newblock \emph{Trends in Food Science \& Technology}, page 104488.

\bibitem[{Maeta et~al.(2015)Maeta, Sasada, and Mori}]{maeta2015framework}
Hirokuni Maeta, Tetsuro Sasada, and Shinsuke Mori. 2015.
\newblock A framework for procedural text understanding.
\newblock In \emph{Proceedings of the 14th International Conference on Parsing Technologies}, pages 50--60.

\bibitem[{Margherita et~al.(2021)Margherita, Elia, and Klein}]{margherita2021managing}
Alessandro Margherita, Gianluca Elia, and Mark Klein. 2021.
\newblock Managing the {C}{O}{V}{I}{D}-19 emergency: A coordination framework to enhance response practices and actions.
\newblock \emph{Technological Forecasting and Social Change}, 166:120656.

\bibitem[{Mar{\i}n et~al.(2021)Mar{\i}n, Biswas, Ofli, Hynes, Salvador, Aytar, Weber, and Torralba}]{marin2021recipe1m+}
Javier Mar{\i}n, Aritro Biswas, Ferda Ofli, Nicholas Hynes, Amaia Salvador, Yusuf Aytar, Ingmar Weber, and Antonio Torralba. 2021.
\newblock Recipe1{M}+: A dataset for learning cross-modal embeddings for cooking recipes and food images.
\newblock \emph{IEEE Transactions on Pattern Analysis and Machine Intelligence}, 43(1):187--203.

\bibitem[{Mehrotra et~al.(2024)Mehrotra, Parab, and Gulwani}]{mehrotra2024enhancing}
Pronita Mehrotra, Aishni Parab, and Sumit Gulwani. 2024.
\newblock Enhancing creativity in large language models through associative thinking strategies.
\newblock \emph{arXiv preprint arXiv:2405.06715v1}.

\bibitem[{Mirowski et~al.(2023)Mirowski, Mathewson, Pittman, and Evans}]{mirowski2023co}
Piotr Mirowski, Kory~W. Mathewson, Jaylen Pittman, and Richard Evans. 2023.
\newblock Co-writing screenplays and theatre scripts with language models: Evaluation by industry professionals.
\newblock In \emph{Proceedings of the 2023 CHI conference on human factors in computing systems}, pages 1--34.

\bibitem[{Mizrahi et~al.(2024)Mizrahi, Kaplan, Malkin, Dror, Shahaf, and Stanovsky}]{mizrahi2024state}
Moran Mizrahi, Guy Kaplan, Dan Malkin, Rotem Dror, Dafna Shahaf, and Gabriel Stanovsky. 2024.
\newblock State of what art? a call for multi-prompt llm evaluation.
\newblock \emph{Transactions of the Association for Computational Linguistics}, 12:933--949.

\bibitem[{Mizrahi and Shahaf(2021)}]{mizrahi202150}
Moran Mizrahi and Dafna Shahaf. 2021.
\newblock 50 ways to bake a cookie: Mapping the landscape of procedural texts.
\newblock In \emph{Proceedings of the 30th ACM International Conference on Information \& Knowledge Management}, pages 1304--1314.

\bibitem[{Morris et~al.(2012)Morris, Burton, Bodily, and Ventura}]{morris2012soup}
Richard~G. Morris, Scott~H. Burton, Paul~M. Bodily, and Dan Ventura. 2012.
\newblock Soup over bean of pure joy: Culinary ruminations of an artificial chef.
\newblock In \emph{ICCC}, pages 119--125. Citeseer.

\bibitem[{Mumford(2003)}]{mumford2003have}
Michael~D. Mumford. 2003.
\newblock Where have we been, where are we going? taking stock in creativity research.
\newblock \emph{Creativity research journal}, 15(2-3):107--120.

\bibitem[{Palomo-Vadillo et~al.(2025)Palomo-Vadillo, Ortega-Larrea, Bordonado-Bermejo, and De-Pablos-Heredero}]{palomo2025developing}
Maite Palomo-Vadillo, Ana-Lucia Ortega-Larrea, Mar{\'\i}a-Julia Bordonado-Bermejo, and Carmen De-Pablos-Heredero. 2025.
\newblock Developing an index for measuring gender lens investing in organizations: the {G}{L}{I}{M}{E}{T}{R}{I}{C}{S} framework.
\newblock \emph{Frontiers in Psychology}, 16:1534355.

\bibitem[{Pan et~al.(2024)Pan, Luo, Wang, Chen, Wang, and Wu}]{pan2024unifying}
Shirui Pan, Linhao Luo, Yufei Wang, Chen Chen, Jiapu Wang, and Xindong Wu. 2024.
\newblock Unifying large language models and knowledge graphs: A roadmap.
\newblock \emph{IEEE Transactions on Knowledge and Data Engineering}, 36(7):3580--3599.

\bibitem[{Peeperkorn et~al.(2024)Peeperkorn, Kouwenhoven, Brown, and Jordanous}]{peeperkorn2024temperature}
Max Peeperkorn, Tom Kouwenhoven, Dan Brown, and Anna Jordanous. 2024.
\newblock Is temperature the creativity parameter of large language models?
\newblock In \emph{ICCC}.

\bibitem[{Pereira and Cardoso(2006)}]{pereira2006experiments}
Francisco~C. Pereira and Am{\'\i}lcar Cardoso. 2006.
\newblock Experiments with free concept generation in {D}ivago.
\newblock \emph{Knowledge-Based Systems}, 19(7):459--470.

\bibitem[{Ramos(2003)}]{ramos2003using}
Juan Ramos. 2003.
\newblock Using {T}{F}-{I}{D}{F} to determine word relevance in document queries.
\newblock In \emph{Proceedings of the first instructional conference on machine learning}, volume 242, pages 29--48. Citeseer.

\bibitem[{Rana and Cheok(2025)}]{rana2025generative}
Sharif Uddin~Ahmed Rana and Adrian~David Cheok. 2025.
\newblock Generative innovation: Leveraging the power of large language models for brainstorming.
\newblock In \emph{The Economics of Talent Management and Human Capital}, pages 175--192. IGI Global.

\bibitem[{Ravi et~al.(1994)Ravi, Rosenkrantz, and Tayi}]{ravi1994heuristic}
Sekharipuram~S. Ravi, Daniel~J. Rosenkrantz, and Giri~Kumar Tayi. 1994.
\newblock Heuristic and special case algorithms for dispersion problems.
\newblock \emph{Operations Research}, 42(2):299--310.

\bibitem[{Rico-Juan and Mic{\'o}(2003)}]{rico2003comparison}
Juan~Ram{\'o}n Rico-Juan and Luisa Mic{\'o}. 2003.
\newblock Comparison of {A}{E}{S}{A} and {L}{A}{E}{S}{A} search algorithms using string and tree-edit-distances.
\newblock \emph{Pattern Recognition Letters}, 24(9-10):1417--1426.

\bibitem[{Ritchie(2007)}]{ritchie2007some}
Graeme Ritchie. 2007.
\newblock Some empirical criteria for attributing creativity to a computer program.
\newblock \emph{Minds and Machines}, 17:67--99.

\bibitem[{Said-Metwaly et~al.(2017)Said-Metwaly, Van~den Noortgate, and Kyndt}]{said2017approaches}
Sameh Said-Metwaly, Wim Van~den Noortgate, and Eva Kyndt. 2017.
\newblock Approaches to measuring creativity: A systematic literature review.
\newblock \emph{Creativity: theories-research-applications.-Warsaw, Poland, 2014, currens}, 4(2):238--275.

\bibitem[{Sarkar and Cooper(2021)}]{sarkar2021generating}
Anurag Sarkar and Seth Cooper. 2021.
\newblock Generating and blending game levels via quality-diversity in the latent space of a variational autoencoder.
\newblock In \emph{Proceedings of the 16th International Conference on the Foundations of Digital Games}, pages 1--11.

\bibitem[{Sawyer and Henriksen(2024)}]{sawyer2024explaining}
Keith~R. Sawyer and Danah Henriksen. 2024.
\newblock \emph{Explaining Creativity: The Science of Human Innovation}.
\newblock Oxford university press.

\bibitem[{Sclar et~al.(2023)Sclar, Choi, Tsvetkov, and Suhr}]{Sclar2023QuantifyingLM}
Melanie Sclar, Yejin Choi, Yulia Tsvetkov, and Alane Suhr. 2023.
\newblock Quantifying language models' sensitivity to spurious features in prompt design or: How {I} learned to start worrying about prompt formatting.
\newblock In \emph{The Twelfth International Conference on Learning Representations}.

\bibitem[{Summers-Stay et~al.(2023)Summers-Stay, Voss, and Lukin}]{summers2023brainstorm}
Douglas Summers-Stay, Clare~R. Voss, and Stephanie~M. Lukin. 2023.
\newblock Brainstorm, then select: A generative language model improves its creativity score.
\newblock In \emph{The AAAI-23 Workshop on Creative AI Across Modalities}.

\bibitem[{Sun et~al.(2023)Sun, Xu, Tang, Wang, Lin, Gong, Ni, Shum, and Guo}]{sun2023think}
Jiashuo Sun, Chengjin Xu, Lumingyuan Tang, Saizhuo Wang, Chen Lin, Yeyun Gong, Lionel Ni, Heung-Yeung Shum, and Jian Guo. 2023.
\newblock Think-on-graph: Deep and responsible reasoning of large language model on knowledge graph.
\newblock In \emph{The Twelfth International Conference on Learning Representations}.

\bibitem[{Thomas et~al.(2025)Thomas, Yee, Mayne, Mathur, Jurafsky, and Gligori{\'c}}]{thomascan2025}
Anna Thomas, Adam Yee, Andrew Mayne, Maya~B Mathur, Dan Jurafsky, and Kristina Gligori{\'c}. 2025.
\newblock What can large language models do for sustainable food?
\newblock In \emph{Forty-second International Conference on Machine Learning}.

\bibitem[{Tian et~al.(2024{\natexlab{a}})Tian, Xia, and Song}]{tian2024largel}
Yuanhe Tian, Fei Xia, and Yan Song. 2024{\natexlab{a}}.
\newblock Large language models are no longer shallow parsers.
\newblock In \emph{Proceedings of the 62nd Annual Meeting of the Association for Computational Linguistics (Volume 1: Long Papers)}, pages 7131--7142.

\bibitem[{Tian et~al.(2024{\natexlab{b}})Tian, Huang, Liu, Jiang, Spangher, Chen, May, and Peng}]{tian2024large}
Yufei Tian, Tenghao Huang, Miri Liu, Derek Jiang, Alexander Spangher, Muhao Chen, Jonathan May, and Nanyun Peng. 2024{\natexlab{b}}.
\newblock Are large language models capable of generating human-level narratives?
\newblock In \emph{Proceedings of the 2024 Conference on Empirical Methods in Natural Language Processing}, pages 17659--17681.

\bibitem[{Utterback(1996)}]{utterback1996mastering}
James~M. Utterback. 1996.
\newblock \emph{Mastering the Dynamics of Innovation}.
\newblock Harvard Business School Press.

\bibitem[{Valls-Vargas et~al.(2017)Valls-Vargas, Zhu, and Ontan{\'o}n}]{valls2017towards}
Josep Valls-Vargas, Jichen Zhu, and Santiago Ontan{\'o}n. 2017.
\newblock Towards automatically extracting story graphs from natural language stories.
\newblock In \emph{AAAI Workshops}.

\bibitem[{Varshney et~al.(2019)Varshney, Pinel, Varshney, Bhattacharjya, Sch{\"o}rgendorfer, and Chee}]{varshney2019big}
Lav~R. Varshney, Florian Pinel, Kush~R. Varshney, Debarun Bhattacharjya, Angela Sch{\"o}rgendorfer, and Y-M Chee. 2019.
\newblock A big data approach to computational creativity: The curious case of chef watson.
\newblock \emph{IBM Journal of Research and Development}, 63(1):7--1.

\bibitem[{Veale and Cardoso(2019)}]{veale2019computational}
Tony Veale and Am{\'\i}lcar Cardoso. 2019.
\newblock \emph{Computational Creativity: The Philosophy and Engineering of Autonomously Creative Systems}.
\newblock Springer.

\bibitem[{Voronov et~al.(2024)Voronov, Wolf, and Ryabinin}]{voronov2024mind}
Anton Voronov, Lena Wolf, and Max Ryabinin. 2024.
\newblock Mind your format: Towards consistent evaluation of in-context learning improvements.
\newblock In \emph{Findings of the Association for Computational Linguistics ACL 2024}, pages 6287--6310.

\bibitem[{Wan et~al.(2024)Wan, Hu, Zhang, Wang, Wen, and Lu}]{wan2024felt}
Qian Wan, Siying Hu, Yu~Zhang, Piaohong Wang, Bo~Wen, and Zhicong Lu. 2024.
\newblock ``{I}t felt like having a second mind'': Investigating human-{A}{I} co-creativity in prewriting with large language models.
\newblock \emph{Proceedings of the ACM on Human-Computer Interaction}, 8(CSCW1):1--26.

\bibitem[{Wang et~al.(2024{\natexlab{a}})Wang, Huang, Shen, and Uzzi}]{wang2024preliminary}
Dawei Wang, Difang Huang, Haipeng Shen, and Brian Uzzi. 2024{\natexlab{a}}.
\newblock A preliminary, large-scale evaluation of the collaborative potential of human and machine creativity.
\newblock \emph{PsyArXiv September, 28.}

\bibitem[{Wang et~al.(2024{\natexlab{b}})Wang, Lipka, Rossi, Siu, Zhang, and Derr}]{wang2024knowledge}
Yu~Wang, Nedim Lipka, Ryan~A Rossi, Alexa Siu, Ruiyi Zhang, and Tyler Derr. 2024{\natexlab{b}}.
\newblock Knowledge graph prompting for multi-document question answering.
\newblock In \emph{Proceedings of the AAAI Conference on Artificial Intelligence}, volume~38, pages 19206--19214.

\bibitem[{Wu et~al.(2025)Wu, Weber, and M{\"u}ller}]{wu2025one}
Zhikun Wu, Thomas Weber, and Florian M{\"u}ller. 2025.
\newblock One does not simply meme alone: Evaluating co-creativity between {L}{L}{M}s and humans in the generation of humor.
\newblock In \emph{Proceedings of the 30th International Conference on Intelligent User Interfaces}, pages 1082--1092.

\bibitem[{Yang et~al.(2023)Yang, Ishay, and Lee}]{yang2023coupling}
Zhun Yang, Adam Ishay, and Joohyung Lee. 2023.
\newblock Coupling large language models with logic programming for robust and general reasoning from text.
\newblock In \emph{Findings of the Association for Computational Linguistics: ACL 2023}, pages 5186--5219.

\bibitem[{Yee-King(2022)}]{yee2022latent}
Matthew Yee-King. 2022.
\newblock Latent spaces: A creative approach.
\newblock In \emph{The Language of Creative AI: Practices, Aesthetics and Structures}, pages 137--154. Springer.

\bibitem[{Yuan et~al.(2022)Yuan, Coenen, Reif, and Ippolito}]{yuan2022wordcraft}
Ann Yuan, Andy Coenen, Emily Reif, and Daphne Ippolito. 2022.
\newblock Wordcraft: Story writing with large language models.
\newblock In \emph{Proceedings of the 27th International Conference on Intelligent User Interfaces}, pages 841--852.

\bibitem[{Yuan and F{\"a}rber(2025)}]{yuan2025hallucinations}
Shuzhou Yuan and Michael F{\"a}rber. 2025.
\newblock Hallucinations can improve large language models in drug discovery.
\newblock \emph{arXiv preprint arXiv:2501.13824v2}.

\bibitem[{Zelikman et~al.(2023)Zelikman, Huang, Poesia, Goodman, and Haber}]{zelikman2023parsel}
Eric Zelikman, Qian Huang, Gabriel Poesia, Noah Goodman, and Nick Haber. 2023.
\newblock Parsel: Algorithmic reasoning with language models by composing decompositions.
\newblock \emph{Advances in Neural Information Processing Systems}, 36:31466--31523.

\bibitem[{Zhang et~al.(2025)Zhang, Wang, Wu, Huang, Wu, Chen, Song, Zhang, Rao, and Yu}]{zhang2025sr}
Jiahuan Zhang, Tianheng Wang, Hanqing Wu, Ziyi Huang, Yulong Wu, Dongbai Chen, Linfeng Song, Yue Zhang, Guozheng Rao, and Kaicheng Yu. 2025.
\newblock {S}{R}-{L}{L}{M}: Rethinking the structured representation in large language model.
\newblock \emph{arXiv preprint arXiv:2502.14352v1}.

\bibitem[{Zhang and Shasha(1989)}]{zhang1989simple}
Kaizhong Zhang and Dennis Shasha. 1989.
\newblock Simple fast algorithms for the editing distance between trees and related problems.
\newblock \emph{SIAM journal on computing}, 18(6):1245--1262.

\bibitem[{Zhang et~al.(2024)Zhang, Peng, and Bollegala}]{zhang2024improving}
Tianhui Zhang, Bei Peng, and Danushka Bollegala. 2024.
\newblock Improving diversity of commonsense generation by large language models via in-context learning.
\newblock In \emph{Findings of the Association for Computational Linguistics: EMNLP 2024}, pages 9226--9242.

\bibitem[{Zhang et~al.(2019)Zhang, Yang, Yuan, Shen, and Carin}]{zhang2019syntax}
Xinyuan Zhang, Yi~Yang, Siyang Yuan, Dinghan Shen, and Lawrence Carin. 2019.
\newblock Syntax-infused variational autoencoder for text generation.
\newblock In \emph{Proceedings of the 57th Annual Meeting of the Association for Computational Linguistics}, pages 2069--2078.

\bibitem[{Zhao et~al.(2023)Zhao, Ji, Zhang, He, Wang, Wang, Feng, and Zhang}]{zhao2023large}
Bowen Zhao, Changkai Ji, Yuejie Zhang, Wen He, Yingwen Wang, Qing Wang, Rui Feng, and Xiaobo Zhang. 2023.
\newblock Large language models are complex table parsers.
\newblock In \emph{Proceedings of the 2023 Conference on Empirical Methods in Natural Language Processing}, pages 14786--14802.

\bibitem[{Zhao et~al.(2025)Zhao, Zhang, Li, and Li}]{zhao2024assessing}
Yunpu Zhao, Rui Zhang, Wenyi Li, and Ling Li. 2025.
\newblock Assessing and understanding creativity in large language models.
\newblock \emph{Machine Intelligence Research}, 22(3):417--436.

\bibitem[{Zhou et~al.(2024)Zhou, Min, Fu, Jin, Huang, Li, Mei, and Jiang}]{zhou2024foodsky}
Pengfei Zhou, Weiqing Min, Chaoran Fu, Ying Jin, Mingyu Huang, Xiangyang Li, Shuhuan Mei, and Shuqiang Jiang. 2024.
\newblock Foodsky: A food-oriented large language model that passes the chef and dietetic examination.
\newblock \emph{arXiv preprint arXiv:2406.10261v1}.

\bibitem[{Zhou et~al.(2025)Zhou, Shen, and Wang}]{zhou2025freeblend}
Yufan Zhou, Haoyu Shen, and Huan Wang. 2025.
\newblock Free{B}lend: Advancing concept blending with staged feedback-driven interpolation diffusion.
\newblock \emph{arXiv preprint arXiv:2502.05606v2}.

\end{thebibliography}
\bibliographystyle{acl_natbib}
\clearpage

\appendix
% \section{Appendix}
\label{sec:appendix}
\section{Fine-Tuned Sentence-BERT Model}
\label{algo:sbert_finetuned}

Our initial experiments with the standard sentence-level Sentence-BERT (SBERT) model revealed that it tends to group recipes %as highly similar {simply} due to their shared format rather than meaningful culinary similarities. Moreover, it placed excessive 
emphasizing  textual instructions while overlooking ingredients. As a result, it fails to distinguish broad categories (e.g., salad vs. soup) and struggles with finer-grained distinctions (e.g., carrot cake vs. cheesecake). %To address these limitations, we fine-tuned SBERT on recipe pairs labeled with similarity scores that incorporate both textual and ingredient-based similarity ). \gabis{I don't like to mix in anecdotal results in the model section (``we tried something and it didn't work'' without any quantitative results), I would remove the part about the vanilla SBERT, or report it with results in the results section.}

To better handle recipe similarity, we fine-tuned a Sentence-BERT model.\footnote{all-distilroberta-v1.} Our fine-tuning dataset consisted of 30K pairs of recipes with their new similarity scores, which equally weighted the original Sentence-BERT score and a Ruzicka-based similarity of the two recipes’ ingredient lists as computed by \citet{mizrahi202150}. The dataset was divided into three equal subsets: (1) 10K pairs of recipes, each pair representing instances of the same dish, (2) 10K pairs of recipes from different dishes within the same category (e.g., carrot cake and cheesecake), and (3) 10K pairs of recipes from entirely different categories (e.g., a dessert and a salad).

To preserve the original model’s capabilities and avoid overfitting, we fine-tuned it for one epoch, reserving 5\% of the pairs for validation, achieving a validation accuracy of 92–95\%. Following Sentence-BERT fine-tuning guidelines, we employed 10\% warm-up steps, used cosine similarity loss as our loss function, and set the maximum sequence length to 512 tokens.

\begin{figure}[t!]
\includegraphics[width=\linewidth]{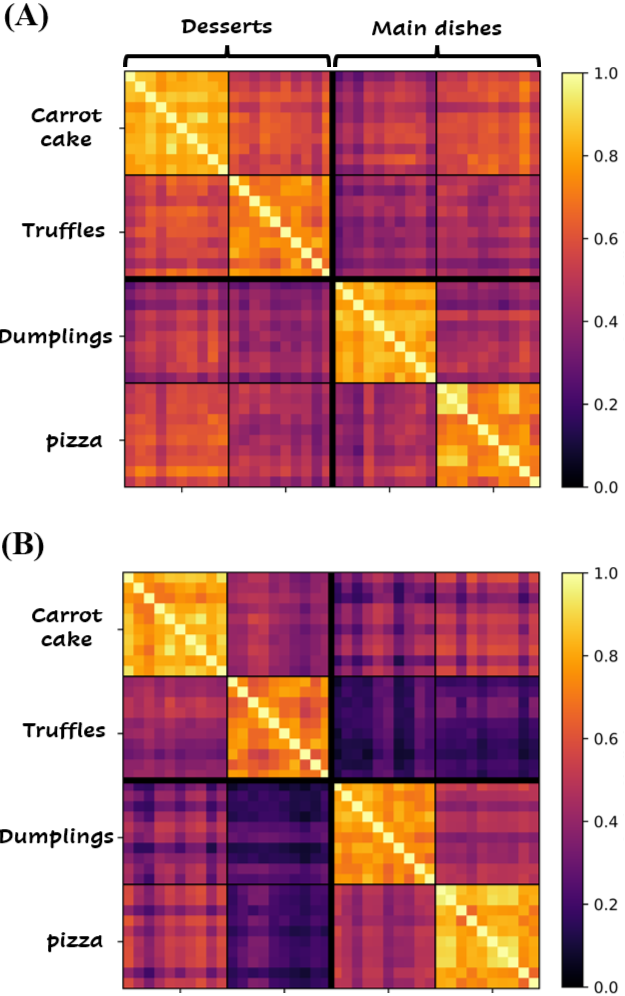}
\caption{\label{fig:sbert_before_after}
Cosine similarity heatmaps of 40 recipes before and after fine-tuning: (A) Sentence-BERT before fine-tuning, (B) Sentence-BERT after fine-tuning. The original Sentence-BERT model shows uniformly high similarity across all dish pairs, while the fine-tuned model maintains high similarity within the same dish and improves differentiation across categories.}\end{figure}

This fine-tuning procedure allowed us to obtain more accurate recipe embeddings that account for both textual instructions and ingredient overlap. Figure~\ref{fig:sbert_before_after} presents a cosine similarity heatmap of 40 recipes before and after fine-tuning: 10 highly similar recipes for Carrot Cake, 10 for Truffles (desserts), 10 for Dumplings, and 10 for Pizza (main dishes). As shown, the original Sentence-BERT model exhibits high similarity scores across all dish pairs, including an unexpectedly high similarity between Carrot Cake and Pizza recipes. In contrast, the fine-tuned model retains high similarity scores within the same dish while improving differentiation between dishes within the same category and across categories.

\section{Text to Tree Parser Details}
\label{app:text_to_tree_details}

\paragraph{Parsing the Ingredients.}
In the first subtask, we instructed the model to parse the ingredients. For each ingredient line, the model extracted (1) the ingredient name, (2) whether the ingredient contributes to the dish’s structural core (e.g., lasagna sheets in lasagna) or its flavor (e.g., lemon in lemon pie), and (3) a simplified base form of the ingredient (e.g., “basil” → “herb,” “walnut” → “nut”).
To evaluate accuracy, we sampled 200 random ingredient lines, and one of us checked whether the parsed information matched the ground truth. We obtained 95\% accuracy for ingredient name parsing, 97.5\% for the structural vs. flavor reference, and 96\% for ingredient base form conversion. These results suggest that \cgpt{} performs reliably in parsing ingredients.

\paragraph{Simplifying the Instructions.}
Next, we asked the model to simplify each sentence in a recipe’s instructions while preserving essential content. Specifically, we instructed it to remove details about quantities, sizes, and descriptive elements, ensuring that each simplified sentence contained exactly one action (placed at the beginning) and that ambiguous instructions were converted into active forms (e.g., “bring to a boil” → “boil”). For example, the instruction “Sprinkle salt over the basil and mozzarella and return to the broiler for 1 to 2 minutes, until the cheese is melted and bubbling.” was simplified into “Sprinkle salt over basil, mozzarella. Broil until cheese melts”. We tested this phase on 50 sampled recipes, resulting in 451 instruction sentences. One of us reviewed these sentences and found that 423 (93.79\%) were simplified correctly. Common errors included using non-informative verbs, omitting a verb or ingredient, or accidentally merging two actions into a single sentence.

\paragraph{Translating the Recipe into a Tree.}
Finally, we used \cgpt{}’s coding proficiency to produce a directed tree representation in DOT, a graph description language. We provided the model with a one-shot example that included a simplified recipe text (dish name, a short list of ingredients, and simplified instructions) along with its corresponding DOT code, annotated with comments to guide the model in structuring the tree representation. To refine the output, we implemented a correction step, automatically removing problematic edges and instructing the model to reconsider them.

\section{Text to Tree Parser Prompts}
\label{app:text_to_tree_prompts}

% We divided the translation process of text to trees into three subtasks using a chain-of-thought approach, as detailed below.

This section includes the prompts we used to prompt \cgpt{} to parse a recipe text into a tree representation. The system message for all prompts presented here was: ``You are a cooking recipe parser''.

% \subsection{Parsing the Ingredients}
\paragraph{Parsing the Ingredients Prompt:}
Given a recipe title, id, and ingredients, for each ingredient, determine: (1) Abbreviation: The shortest clear description. (2) Reference Type: Identify if the ingredient is for structure (`structure') or taste (`taste') of the dish. Ingredients impacting both are labeled as 'taste'. (3) Core Ingredient: Boolean indicating if the ingredient is essential to the identity of the dish (e.g., True for chocolate in chocolate cake). (4) Abstraction: Simplify the ingredient to its base form (e.g., `basil' to `herb', `walnuts’ to `nut’, `eggs' to `egg'). Please return the results in the following JSON format only: \{``recipe\_id'': [(abbreviation, ref, core, abstraction), ...], ...\}. INPUT: recipe\_title, recipe\_id, ingr\_list
\textbackslash n 
recipe\_title, recipe\_id, ingr\_list
\textbackslash n 
... OUTPUT:

% \subsection{Simplifying the Instructions}
\paragraph{Simplifying the Instructions Prompt:}
Given the following cooking instructions, please simplify and shorten them as much as possible. Remove quantities, sizes, and descriptions. Ensure each verb initiates a new sentence, and that a sentence does not contain two verbs. Convert permissive or ambiguous instructions into active forms (e.g., ``let cool'' -> ``cool'', ``alternate layers'' -> ``layer''). Return output in JSON format with the key as `recipe\_id' and the value as the full simplified text. INPUT: \{recipe\_id: <instruction text>, ...\} OUTPUT:

% \subsection{Translating a Single Recipe into Tree (1-shot)}
\paragraph{Translating the Recipe into a Tree Prompt:}
Title: ... Ingredients: ... Directions: ... Code: ...  \# end of code (the 1-shot example)
\\
Title: <dish\_name> Ingredients: <ingrendient\_abbreviation\_list> 
Directions: [i1] <1st\_direction> [i2] <2nd\_direction> ... Code: 

% \subsection{Tree Correction}
\paragraph{Tree Correction Prompt:}
You are provided with the title, ingredients, and directions of a recipe, along with a partial Dot code that represents the recipe's tree structure. The Dot code is missing some edges. Additionally, you will receive names of nodes for which these connections are missing. For each provided node name, add exactly one edge from this node to the action node that uses it (if it is an ingredient) or processes its outcome (if it is an action). Please return only the Dot code for these specific edges, including necessary comments, and exclude any additional text.
Title: <dish\_name>
Ingredients: <ingredient\_abbreviation\_list>
Directions:
\texttt{[i1]} <1st\_direction>
\texttt{[i2]} <2nd\_direction>
...
Partial Dot code:
<dot\_code>
Name of nodes with missing edges:
<node\_names>
OUTPUT:

\section{Tree Edit Distance Implementation}
\label{app:tree_edit_distance}

In this appendix, we describe our approach for computing the minimal edit distance between recipe trees. As mentioned in Section~\ref{algo:edit_distance}, we employ the Zhang–Shasha algorithm \cite{zhang1989simple}, which extends the well-known string edit distance approach to ordered labeled trees. Specifically, a \textit{\textbf{labeled tree}} is one in which each node is assigned a symbol from a fixed finite alphabet, and an \textit{\textbf{ordered tree}} is one in which each set of siblings has a defined left-to-right order. While computing tree edit distance for unordered trees is NP-hard and even MAX SNP-hard \cite{arora1998proof}, the Zhang–Shasha algorithm provides a polynomial-time solution for the ordered case. To make our labeled recipe trees compatible with this approach, we impose an ordering on sibling nodes by sorting them lexicographically according to their labels.

We further adjust edit costs to encourage matching analogous nodes across recipes. Specifically, we allow a node to be substituted by another only if both nodes share the same type (i.e., both are ingredient nodes or both are action nodes). Moreover, a substitution between two nodes 
% with a similar 
incurs zero cost if they have the same label, or a small fixed cost if their labels share the same abstract meaning. For example, two ingredients both categorized as “herb” may replace one another at a lower cost than an “herb” and a “liquid” ingredients. Similarly, two action nodes categorized under “heat application” are more likely to substitute for each other than an action node categorized under “heat application” and another one that is categorized under “flavor enhancement”.

To determine whether two ingredient nodes share the same abstraction, we use the ingredient abstraction obtained from parsing the ingredients (see Appendix~\ref{app:text_to_tree_details}). To determine whether two action nodes share the same abstraction (e.g., “heat application” for “bake” and “microwave”), we collected the 250 most common action verbs in recipes and created a hierarchy, grouping these verbs into categories such as heat application, preparation, positioning, flavor enhancement, etc. 

Formally, let $T_1$ and $T_2$ be two recipe trees composed of ingredient and action nodes. We allow the operations of insertion, deletion, and update. The cost of insertion or deletion is set to 100, whereas the update cost depends on whether the two nodes share the same type and the same label or abstraction. If they have the same type and the same label, the cost is 0; if they share the same type but only the same abstraction, it is 5; otherwise, the cost is $\infty$, indicating an infeasible substitution.
This cost scheme encourages the edit distance algorithm to favor substituting analogous parts over insertions and deletions, resulting in more semantically meaningful transformations.

As noted in Section~\ref{algo:edit_distance}, 
stopping at different points in the transformation process can create unique dishes (see Figure~\ref{fig:edit_distance_process}). Additionally, shuffling the order of edits can generate entirely new intermediate ideas. 
% To avoid selecting a variant too similar to either the source or target recipe, we exclude the first and last quarters of the possible intermediate transformations and select from those in the middle.
To produce more coherent results, we impose a partial order on the shuffled operations. We prioritize inserting and updating key flavor ingredients from the target recipe (e.g., “lemon” in a lemon pie) so they appear earlier in the transformation. At the same time, we delay deleting or updating structural ingredients from the source recipe (e.g., “lasagna sheets” in lasagna) to preserve its core structure. We determine which ingredients contribute to flavor and which to structure during the parsing phase (see Appendix~\ref{app:text_to_tree_details}). This approach helps maintain the essential characteristics of both dishes, integrating distinct flavor components from the target while retaining the structural integrity of the source. Notably, reversing the transformation (dish B → dish A) results in a different edit sequence, leading to distinct new recipes.

To ensure the dishes are indeed recombinations, we discard any recipe that lacks at least one \textbf{essential ingredient} from both original dishes (if such an ingredient exists). We define an essential ingredient as one that appears frequently in recipes for a given dish but is not broadly common across all recipes (e.g., lasagna sheets in lasagna).
Additionally, we remove ideas that are too similar to the seed recipes. To ensure that the combined tree preserves cross-dish inspiration, we require that at least 30\% of its elements (nodes and edges) come from each of the original recipes.

\section{Identifying Conflicting Ingredients}
\label{app:conflict}

We remove ingredients that appear to collide in their flavors.  Specifically, we look for pairs of ingredients that seldom appear together in the recipe repository, treating their pairing as uncommon and attempt to determine whether it might be a creative success or a failure.  %\gabis{Unclear if this is done automatically}
To do so, we rely on two external datasets.
First, we use \textbf{flavorDB} \cite{garg2018flavordb}, which catalogs taste molecules for a wide range of raw ingredients such as fruits, vegetables, and fish. Inspired by this dataset owner’s claim that two ingredients pair well if they share a larger proportion of taste molecules, we define a Jaccard-based pairing score between raw ingredients. Since flavorDB does not cover processed ingredients (e.g., lasagna sheets that consist of flour, water, and eggs), we also use \textbf{FoodData Central} \cite{fukagawa2022usda} to infer their raw components. We define the pairing score
between two composite ingredients as the lowest score among their constituent raw-ingredient
pairs. After exploration, we chose 0.3 as our value
threshold. If the score falls below 0.3, we consider the pairing problematic.

% However, since flavorDB only covers raw ingredients, we also incorporate data from \textbf{FoodData Central} \cite{fukagawa2022usda}, which lists the raw components of processed or composite ingredients (e.g., lasagna sheets consist of flour, water, and eggs). 
% This allows us to extend our pairing score to any ingredient combination within a recipe. 
% Specifically, we define the pairing score between two composite ingredients to be the minimum raw-ingredient pairing score among all their constituent raw-ingredient pairs. If the score falls below 0.3, we consider the pairing problematic. To resolve such conflicts, we iteratively remove the ingredient with the highest number of low-scoring pairings until no further collisions remain.

\section{Tree to Text Prompts}
\label{app:tree_to_text_prompts}

This section includes the prompts we used to prompt \cgpt{} to translate structured tree representation back into natural language. The system message for all prompts presented here was: ``You are a cooking expert''.

% \subsection{Translate a tree into a raw recipe}
\paragraph{Translate Tree into Raw Recipe Prompt:}
Given the following DOT code, which represents a recipe graphically by defining ingredient nodes, action nodes, and their interconnections, translate the structure into a natural language recipe. The DOT code maps each ingredient to specific actions, and it outlines the order of these actions to demonstrate the cooking process.
DOT CODE:
```<recipe\_idea\_dot\_code>'''
Convert this structured representation into a detailed cooking recipe in natural language. Requirements: (1) Output should only include the title, ingredients with quantities, and sequential instructions. (2) Avoid any explanatory comments or embellishments.
OUTPUT:

% \subsection{Find Issues and Correct Recipe (two steps)}
\paragraph{Find Issues and Correct Recipe Prompts:}
Step I: Review the recipe provided below, which is written in natural language. Identify and list any potential issues with it, excluding any concerns related to unconventional ingredient combinations. Please provide only a list of potential issues without revising the recipe. 
RECIPE:
```<GENERATED RECIPE>'''
\\
\\[-0.8em]
Step II: Please edit the recipe to address the identified issues. Ensure the recipe remains as a single, unified component. Output only the corrected version of the recipe.
OUTPUT:

\paragraph{Summarize Recipe Prompt:}
Please summarize the following recipe in a few sentences: (1) Start with a super concise description of the dish, focusing *only* on its final result. (2) Then, provide a summary of the recipe, including its main components, actions, and all the ingredients used. Use a descriptive tone for this part, avoiding imperative sentences.
RECIPE:
```<full\_recipe>'''

\paragraph{Review Ingredients Prompt:}
You are given a description of a creative recipe.
CREATIVE RECIPE DESCRIPTION:
```<creative\_recipe\_description>'''
Your task is to preserve the creative ingredients in the recipe while suggesting the removal or substitution of ingredients that might negatively impact the dish's flavor. You should: (1) Recognize the unique and unusual ingredients that contribute to the creativity of the dish. (2) Systematically compare all pairs of ingredients in the dish and identify ingredients that have a clear, strong clash with each other due to conflicting flavors. Be thorough and ensure that you include all possible pairs of ingredients that have a strong clash. (3) Based on the identified strong clashes, suggest removals and substitutions of ingredients to avoid clashes, while preserving the creative aspects of the dish.
Return only the following JSON output format:
\{``dish\_ingredients'': <list of strings: the full list of ingredients in the dish>, ``creative\_ingrs'': <list of strings: the list of ingredients that contribute creatively to the dish>, ``flavor\_clashes'': <list of string pairs: the clashing ingredients>, ``removals'': <list of strings: the list of ingredients to remove>, ``substitutions'': <list of string pairs: ingredients to substitute - (ingr1, ingr2) means `replace ingr1 in ingr2'>\}

\paragraph{Increase Readability Prompt:}
Given the following recipe: (1) Remove the following ingredients: <bad\_ingredients>. (2) Make the following ingredient substitutions: <required\_substitutions>. (3) Split its ingredients and instructions into distinct sections to improve readability (e.g., ``mix dry ingredients'', ``assemble'', etc.). You can change the order of lines but keep the content unchanged.
```<full\_recipe>'''

\section{Chosen Prompts for Experiments}
\label{app:exp_prompts}

In this section, we present the prompts used to guide \cgpt{} in generating recipes for both experiments.

\paragraph{Experiment 1 Prompt:}
% \label{app:exp_prompt1}
You are a chef at a fusion restaurant that excels in creating delightful and unexpected combinations of classic dishes. Your task today is to design an innovative recipe that merges the intricate layers of \{dish1\} with the rich decadence of \{dish2\}. Develop a comprehensive recipe that includes:
(1) A unique name that embodies the essence of this fusion dish.
(2) A detailed list of ingredients.
(3) Step-by-step cooking and assembly instructions, highlighting inventive cooking techniques or unusual ingredient interactions.
Promote bold experimentation with flavors and textures to create a dish that is both surprising and satisfying.
\\
\\ [-0.8em]
Following prompt: Design another different innovative recipe that merges the intricate layers of \{dish1\} with the rich decadence of \{dish2\}.

\paragraph{Experiment 2 Prompt:}
% \label{app:exp_prompt2}
What is the most creative and out-of-the-box recipe you can create?
\\
\\ [-0.8em]
Following prompt: Create a fresh, unique recipe that differs from the previous ones but matches their level of creativity.

\section{Human Experiment Questionnaire}
\label{app:human_exp_questions}
\begin{enumerate}
    \item Do the instructions in this recipe make sense? 
    A recipe that doesn’t make a lot of sense contains technical issues. Issues could be minor (for example, mixing an already-mixed salad) or major (not cooking raw chicken). 
    \{scale (1-5): 1: Throw away the recipe, too many changes needed, 2: Need to change most of the recipe, 3: Requires a lot of changes, 4: Almost perfect, requires some minor changes, 5: Makes perfect sense, could cook it as is\}
    \item Do the combination of ingredients in this recipe make sense? 
    \{scale: same\}
    \item How similar is this recipe to others you have seen or used? 
    \{scale (1-5): 1: Very similar to many recipes, 2: Similar but not very common, 3: Somewhat similar to others, 4: Quite different from most recipes, 5: Very different, highly unusual\}
    \item How novel is the way the instructions are combined in this recipe compared to typical recipes? 
    \{scale: same\}
    \item How novel is the combination of the ingredients in this recipe compared to typical recipes?  
    \{scale: same\}
    \item Assuming a cook followed this recipe, would people want to taste it? 
    \{scale (1-5): 1: Never, 2: Only if they have to, 3: Only if they’re really hungry, 4: They’d probably try it, 5: Yes, definitely\}
    \item Assuming a cook followed this recipe after the required modifications, would people want to taste it?
    \{scale: same\}
    \item How original do you find this recipe overall? (after the required modifications)
    \{scale (1-5): 1: Not original at all, 2: Slightly original, 3: Somewhat original, 4: Fairly original, 5: Extremely original and creative\}
    
\end{enumerate}

\end{document}